%% file: main.tex
\documentclass[sigconf]{acmart}

\copyrightyear{2024}
\acmYear{2024}
\setcopyright{acmlicensed}\acmConference[KDD '24]{Proceedings of the 30th ACM SIGKDD Conference on Knowledge Discovery and Data Mining}{August 25--29, 2024}{Barcelona, Spain}
\acmBooktitle{Proceedings of the 30th ACM SIGKDD Conference on Knowledge Discovery and Data Mining (KDD '24), August 25--29, 2024, Barcelona, Spain}
\acmDOI{10.1145/3637528.3671917}
\acmISBN{979-8-4007-0490-1/24/08}

\usepackage{subfigure}
\usepackage{multirow}
\usepackage{bm}
\usepackage{xcolor}
\usepackage{color, soul}
\definecolor{codegray}{rgb}{0.5,0.5,0.5}

\usepackage{multirow}
\usepackage{enumitem}
\usepackage{graphicx}
\usepackage{algorithmic}
\usepackage[ruled,vlined,linesnumbered]{algorithm2e}
\usepackage{enumitem}

\begin{document}

\title{Graph Condensation for Open-World Graph Learning}

\author{Xinyi Gao}
\orcid{0009-0004-1146-8925}
\affiliation{
\institution{The University of Queensland}
\city{Brisbane}
\country{Australia}
}
\email{xinyi.gao@uq.edu.au}

\author{Tong Chen}
\affiliation{
\institution{The University of Queensland}
\city{Brisbane}
\country{Australia}
}
\email{tong.chen@uq.edu.au}

\author{Wentao Zhang}
\affiliation{
\institution{Peking University}  
\city{Beijing}
\country{China}
}
\email{wentao.zhang@pku.edu.cn}

\author{Yayong Li}
\affiliation{
\institution{}
\institution{Data 61, CSIRO}
\city{Brisbane}
\country{Australia}
}
\email{yayongli@outlook.com}

\author{Xiangguo Sun}
\affiliation{ 
\institution{The Chinese University of Hong Kong}  
\city{Hong Kong}
\country{China}
}
\email{xiangguosun@cuhk.edu.hk}

\author{Hongzhi Yin}
\authornote{Corresponding author.}
\affiliation{
\institution{The University of Queensland}  
\city{Brisbane}
\country{Australia}
}
\email{h.yin1@uq.edu.au}

\renewcommand{\shortauthors}{Xinyi Gao et al.}

\begin{abstract}
The burgeoning volume of graph data presents significant computational challenges in training graph neural networks (GNNs), critically impeding their efficiency in various applications. To tackle this challenge, graph condensation (GC) has emerged as a promising acceleration solution, focusing on the synthesis of a compact yet representative graph for efficiently training GNNs while retaining performance. Despite the potential to promote scalable use of GNNs, existing GC methods are limited to aligning the condensed graph with merely the observed static graph distribution. This limitation significantly restricts the generalization capacity of condensed graphs, particularly in adapting to dynamic distribution changes. In real-world scenarios, however, graphs are dynamic and constantly evolving, with new nodes and edges being continually integrated. Consequently, due to the limited generalization capacity of condensed graphs, applications that employ GC for efficient GNN training end up with sub-optimal GNNs when confronted with evolving graph structures and distributions in dynamic real-world situations. To overcome this issue, we propose open-world graph condensation (OpenGC), a robust GC framework that integrates structure-aware distribution shift to simulate evolving graph patterns and exploit the temporal environments for invariance condensation. This approach is designed to extract temporal invariant patterns from the original graph, thereby enhancing the generalization capabilities of the condensed graph and, subsequently, the GNNs trained on it. Furthermore, to support the periodic re-condensation and expedite condensed graph updating in life-long graph learning, OpenGC reconstructs the sophisticated optimization scheme with kernel ridge regression and non-parametric graph convolution, significantly accelerating the condensation process while ensuring the exact solutions. Extensive experiments on both real-world and synthetic evolving graphs demonstrate that OpenGC outperforms state-of-the-art (SOTA) GC methods in adapting to dynamic changes in open-world graph environments.
\end{abstract}


  
\maketitle

\input{1int}

\input{2pre}
\input{3met}

\input{4exp}
\input{5con}

\input{ack}

\bibliographystyle{ACM-Reference-Format}
\bibliography{ref}

\input{6app}

\end{document}

%% file: 1int.tex
\section{Introduction}
Graph data~\cite{sun2023all,gao2023accelerating,yang2023time} is used to represent complex structural relationships among various entities and has enabled applications across a diverse range of domains, such as chemical molecules \cite{guo2022graph}, social networks \cite{long2023model,li2018influence,jung2012irie,nguyen2017retaining,sun2023self}, and recommender systems \cite{zhang2022pipattack,long2023decentralized,yin2024device}. 
However, the exponential growth of data volume in these applications poses significant challenges in data storage, transmission, and particularly the training of graph neural networks (GNNs) \cite{qu2021imgagn,zheng2016keyword,liu2023imbalanced,gao2023semantic}.
These challenges become more pronounced in scenarios that require training multiple GNN models, such as neural architecture search~\cite{zhang2022pasca}, continual learning~\cite{rebuffi2017icarl, yang2023time}, and federated learning~\cite{pan_fedgkd_2023}.
Consequently, there is a pressing need for more efficient methodologies for processing large-scale graph data.
In response, graph condensation (GC) \cite{gao2024graph,jin2022graph,jin2022condensing,loukas2018spectrally} has emerged, aiming to synthesize a compact (e.g., $1,000\times$ smaller) yet informative graph that captures essential characteristics of the large original graph. 
The condensed graph enables fast training of numerous GNNs bearing different architectures and hyper-parameters, while ensuring their performance is comparable to the ones trained on the original graph.
As such, in the deployment stage, these trained GNNs can directly perform inference on the original graph to support various downstream tasks.

\begin{figure}[t]
\setlength{\abovecaptionskip}{0.1cm}
\centering
\begin{minipage}[t]{0.98\linewidth}
\centering
\includegraphics[width=\linewidth]{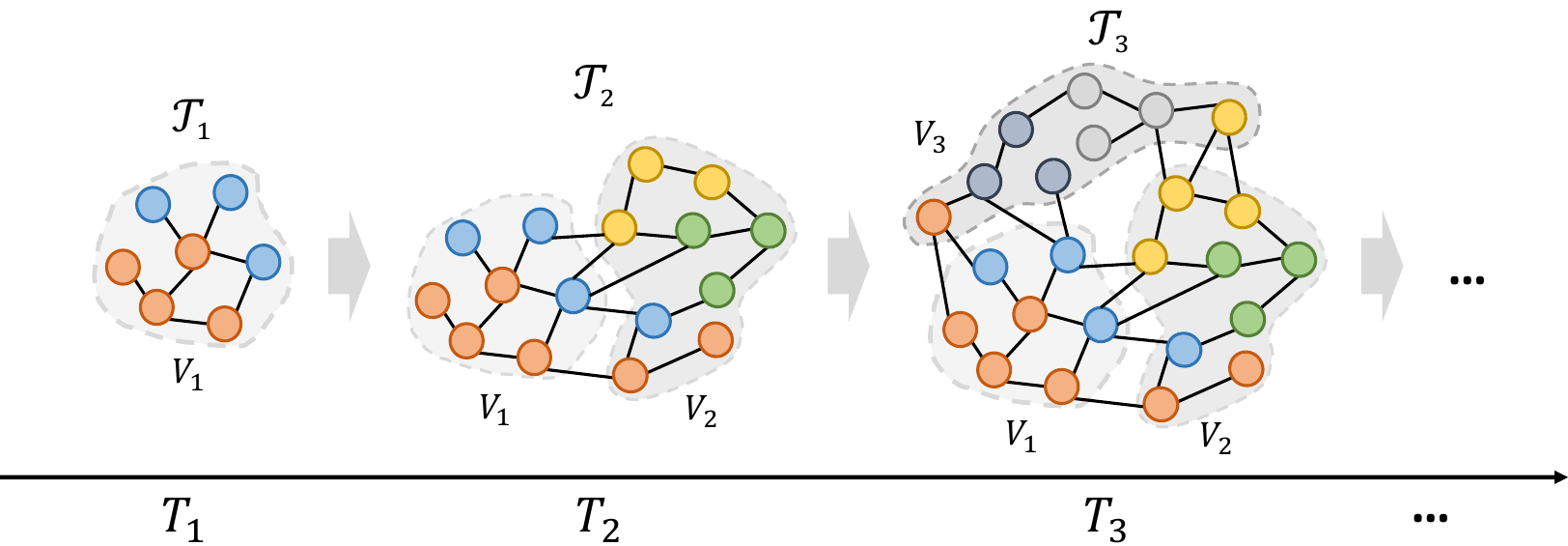}
\end{minipage}
\begin{minipage}[t]{0.49\linewidth}
\centering
\includegraphics[width=\linewidth]{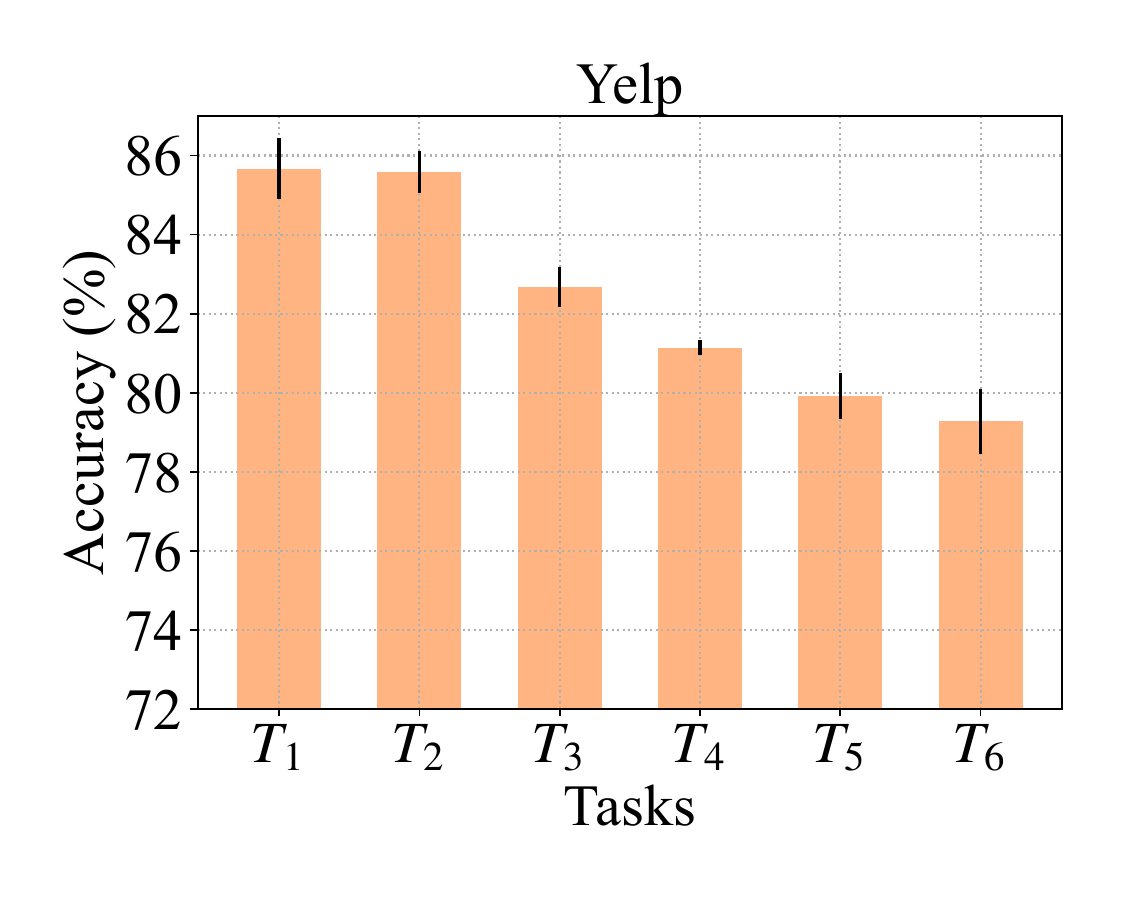}
\end{minipage}
\begin{minipage}[t]{0.49\linewidth}
\centering
\includegraphics[width=\linewidth]{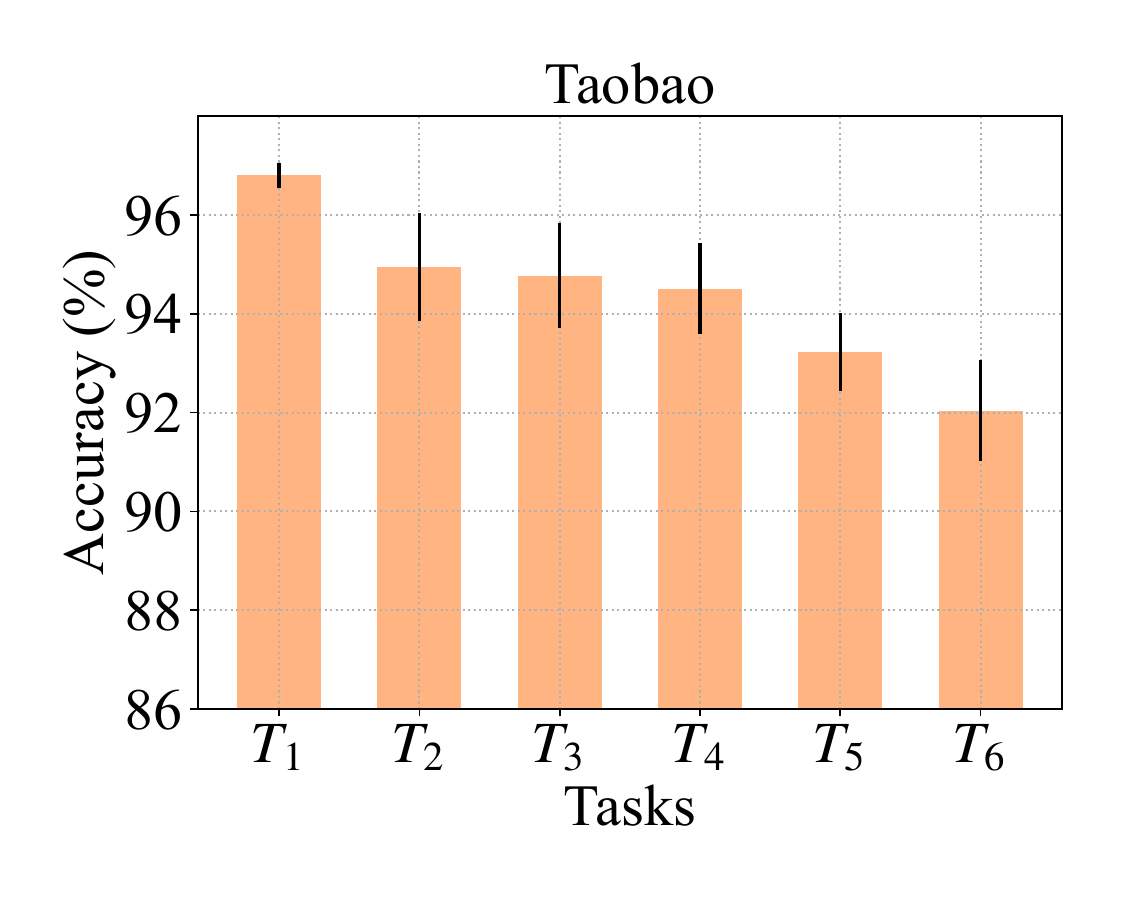}
\end{minipage}
\caption{The upper panel presents the evolution of the graph. The graph $\mathcal{T}_i$ expands as tasks ${T}_i$ progress. Varying colors of nodes represent distinct classes. The lower panel shows the test accuracy of consecutive tasks on the Yelp and Taobao datasets. The test model is GCN, which is trained on the condensed graph of the initial task $T_{1}$ and applied to evaluate subsequent tasks without fine-tuning. The test set expands following the tasks, and the evaluation is limited to the nodes belonging to the classes in $T_{1}$.}
\label{fig_obs}
\end{figure}

Despite the potential to accelerate model training, current GC methods still fall short in their real-world practicality.
By default, conventional GC methods are subsumed under a static setting, which requires that a large graph is firstly condensed to facilitate GNN training on the small graph, and the trained GNNs are then deployed on the same large graph for testing. Unfortunately, assuming the large graph stays unchanged throughout the entire process severely contradicts the dynamic and evolving nature of graph-structured data in the real world.
In fact, graphs in many high throughput applications are inherently \textit{open-world} \cite{geng2020recent,bendale2016towards,wu2020openwgl, feng2023towards}, where new nodes and classes continuously emerge and are integrated into the existing graph structure, as depicted in Figure \ref{fig_obs}.
This phenomenon is exemplified in citation networks \cite{DBLP:conf/nips/HuFZDRLCL20,page1999pagerank}, where some new papers explore established topics, while others venture into emerging areas. These papers in the novel areas are often developed from previous studies and cite a range of related literature. 
Such addition of nodes often introduces novel patterns that are distinctly different from those previously observed, increasing the graph's complexity and diversity. 
Consequently, in addition to preserving performance on the initial large graph, GNNs trained on condensed graphs are expected to exhibit adaptability to novel patterns that emerge within dynamic environments.
For instance, GC for neural architecture search \cite{zhou2022auto,ding_faster_2022} initially condenses a snapshot of the original graph, thereafter employing the condensed graph to accelerate the searching procedure and identify optimal model architecture. However, the evolving nature of graphs leads to a discrepancy between the model deployment environment and the condensed graph snapshot.
Therefore, optimal model architecture identified based on condensed graphs should sustain its superior performance over time, even as the graph evolves.
In a nutshell, the critical problem that arises for GC in open-world scenarios is: ``\textit{How can GC methods be adapted to handle the dynamic nature of evolving graphs, ensuring that GNNs trained on condensed graphs remain accurate and robust in the face of continual changes in graph structures?}''

Specifically, the application of GC within open-world graph environments encounters two primary challenges.
The first challenge arises from the distribution shift caused by the constant addition of new nodes, which may either belong to existing categories or introduce entirely new classes.
On the one hand, these nodes typically exhibit distributions that diverge from existing ones.
On the other hand, their integration, particularly when involving novel classes, can modify the distribution patterns of previously observed nodes through their connections.
However, the condensed graph created by existing GC methods, intended to act as a simple data simulator capturing a static view of the original graph's distribution, inherently constrains the generalization capacities of GNNs trained on these graphs to distribution shifts.
To assess the effect of newly added nodes to GC, we simulate the deployment of GC on two real-world, progressively evolving graphs: Yelp and Taobao\footnote{We construct these datasets according to actual timestamps as detailed in Section \ref{exp_set}.}. 
The GCN \cite{DBLP:conf/iclr/KipfW17} is trained on the condensed graph of the initial task $T_{1}$ and applied to evaluate distinct test sets in subsequent tasks \textit{without} fine-tuning. 
As depicted in Figure \ref{fig_obs}, there was a noticeable decline in the classification performance of test nodes in each task, highlighting the distribution shift caused by newly added nodes and the limited adaptability of GC within dynamic graphs.

In the meantime, the second challenge involves intensive computation in the condensation process. 
The continuous addition of new nodes and subsequent distribution changes necessitate periodic re-condensation to refresh and realign the condensed graph with evolving data distributions. 
However, the condensation process is often complex and slow to converge \cite{jin2022graph}, resulting in a time-consuming procedure that hampers efficient life-long graph data management.
In GC approaches, the process begins by encoding both the large original graph and the condensed graph through a relay model. Subsequently, the relay model, along with the condensed graph, is updated utilizing a nested loop optimization strategy.
This GC paradigm is implemented iteratively until convergence is achieved.
However, the iterative encoding of the large-scale original graph, as well as the sophisticated nested loop optimization, inherently demands substantial time and extensive computational resources. 
Consequently, these intensive computations pose a significant obstacle to the serving of GC in evolving and life-long graph systems.

In light of these challenges, we propose a novel graph condensation approach, \underline{Open}-wold \underline{g}raph \underline{c}ondensation (OpenGC), facilitating trained downstream GNNs to handle evolving graphs in the open-world scenarios.
To tackle the distribution shift issue, we propose temporal invariance condensation that incorporates invariant learning \cite{arjovsky2019invariant} to preserve the invariant patterns across different temporal environments in the condensed graph. 
To this end, a temporal data augmentation technique is designed to simulate the graph's evolving pattern and exploit the structure-aware distribution shift by referring to the historic graph.
It specifically considers the similarity of node pairs and targets the low-degree nodes, which are more susceptible to the influence of newly added neighbors.
By this means, OpenGC can significantly enhance the adaptability of downstream GNNs by training them on the temporally generalized condensed graph, thus eliminating the need to laboriously design specific generalization modules for each GNN.
In response to the existing sophisticated condensation procedure, we design an efficient GC paradigm that combines Kernel Ridge Regression (KRR) with non-parametric graph convolution, circumventing the nested loop optimization and heavy graph kernel encoding in other KRR-based GC methods~\cite{xu2023kernel,wang2023fast}.
Consequently, OpenGC is well-suited for managing life-long graph data and handling the continuous growth and dynamic changes of open-world graphs. The main contributions of this paper are threefold:
\begin{itemize}[leftmargin=*]
\item \textbf{New problem and insights.} We are the first (to the best of our knowledge) to focus on the practical deployment issue of GC in the evolving open-world graph scenario and point out the necessity of learning temporally generalized condensed graphs, which is an important yet under-explored problem in GC.
\item \textbf{New methodology.} We present OpenGC, a GC method that explores the temporal invariance pattern within the original graph, endowing the condensed graph with temporal generalization capabilities. Additionally, OpenGC employs an efficient condensation paradigm, significantly improving the condensation speed and enabling prompt updates to the condensed graph.
\item \textbf{SOTA performance.} 
Through extensive experimentation on both real-world and synthetic evolving graphs, we validate that OpenGC excels in handling distribution shifts and accelerating condensation procedures, surpassing various state-of-the-art (SOTA) GC methods in performance.
\end{itemize}

%% file: 2pre.tex
\section{Preliminaries}
In this section, we first introduce the GNN and conventional GC, then formally define the problem studied.

\subsection{Graph Neural Networks}
\label{sec_gcn}
Consider that we have a large-scale graph $\mathcal{T}=\{{\bf A}, {\bf X}\}$ consisting of $N$ nodes. ${\bf X}\in{\mathbb{R}^{N\times d}}$ denotes the $d$-dimensional node feature matrix and ${\bf A}\in \mathbb{R}^{N\times N}$ is the adjacency matrix. We use ${\bf Y}\in{\mathbb{R}^{N\times C}}$ to denote the one-hot node labels over $C$ classes.
GNNs learn the embedding for each node by leveraging the graph structure information and node features as input.
Without loss of generality, we use graph convolutional network (GCN) \cite{DBLP:conf/iclr/KipfW17} as an example, where the convolution operation in the $k$-th layer is defined as follows: 
\begin{equation}
{\bf{H}}^{(k)} =\text{ReLU}\left(\hat{\mathbf{A}}{\bf H}^{(k-1)}{\mathbf W}^{(k)}\right),   
\label{eq_gcn}
\end{equation}
where ${\bf{H}}^{(k)}$ is the node embeddings of the $k$-th layer, and ${\bf{H}}^{(0)}=\mathbf{X}$.
$\hat{\mathbf{A}}=\widetilde{\mathbf{D}}^{-\frac{1}{2}}\widetilde{\mathbf{A}}\widetilde{\mathbf{D}}^{-\frac{1}{2}}$ is the normalized adjacency matrix. 
$\widetilde{\mathbf{A}}$ represents the adjacency matrix with the self-loop, $\widetilde{\mathbf{D}}$ denotes the degree matrix of $\widetilde{\mathbf{A}}$, and $\mathbf{W}^{(k)}$ is the trainable weights at layer $k$. 
For simplicity, we denote an $K$-layer GNN encoder as $\mathbf{H}=h \left(\mathcal{T} \right)$, where $\mathbf{H}$ denotes the final node embeddings utilized in downstream tasks. 
In this paper, we concentrate on the node classification task, where $\mathbf{H}$ is further input into the classifier $g\left(\cdot\right)$. Consequently, the entire model is denoted as $f = g \circ h$, encapsulating both the GNN encoder and the classifier.

\subsection{Graph Condensation}
\label{sec_gc}
Graph condensation \cite{jin2022graph} aims to generate a small synthetic graph $\mathcal{S}=\{{\bf A'}, {\bf X'}\}$ with ${\bf A'}\in\mathbb{R}^{N'\times N'}$, ${\bf X'}\in\mathbb{R}^{N'\times d}$ as well as its label ${\bf Y'}\in{\mathbb{R}^{N'\times C}}$, where $N'\ll{N}$.
The model trained on $\mathcal{S}$ can achieve comparable node classification performance to the one trained on the much larger $\mathcal{T}$.
To facilitate the connection between real graph $\mathcal{T}$ and synthetic graph $\mathcal{S}$, a relay model $f_{\theta}= g \circ h$ parameterized by ${\theta}$ is employed in the optimization process for encoding both graphs.
We initially define the loss for $\mathcal{T}$ and $\mathcal{S}$ about the parameter ${\theta}$ as:
\begin{equation}
\label{eq_iniloss}
\begin{split}
&\mathcal{L}^{\mathcal{T}} \left( \theta \right) = \ell \left( f_{\theta} \left( {\mathcal{T}} \right) , \mathbf{Y} \right),\\
&\mathcal{L}^{\mathcal{S}} \left( \theta \right)  = \ell \left( f_{\theta} \left( {\mathcal{S}} \right) , \mathbf{Y}' \right),  
\end{split}
\end{equation}
where $\ell$ is the classification loss such as cross-entropy and $\mathbf{Y}'$ is predefined to match the class distribution in $\mathbf{Y}$.
Then the objective of GC can be formulated as a bi-level optimization problem:
\begin{equation}
\label{eq_oriloss}
\min_{\mathcal{S}} \mathcal{L}^{\mathcal{T}}  \left( \theta^{\mathcal{S}} \right)
\:\: \text{s.t.}  \:\: \theta^{\mathcal{S}}   = \arg\min_{\mathcal{\theta}} \mathcal{L}^{\mathcal{S}}  \left( \theta \right) .
\end{equation}
To solve the objective outlined above, conventional GC~\cite{jin2022graph} proposes to match the model gradients at each training step $t$. In this way, the training trajectory on condensed data can mimic that on the original data, i.e., the models trained on these two datasets converge to similar solutions.
\begin{equation}
\label{eq_gmloss}
\begin{split}
& \min_{\mathcal{S}} \mathbb{E}_{\theta_0 \sim {\Theta}} \left[ \sum_{t=1}^{T} \mathcal{D} \left( \nabla_{\theta_t}\mathcal{L}^{\mathcal{T}} \left( {\theta_t}\right), \nabla_{\theta_t}\mathcal{L}^{\mathcal{S}} \left( {\theta_t} \right) \right) \right] \,\, \\
& \text{s.t.} \,\, {\theta}_{t+1} = \operatorname{opt}\left( \mathcal{L}^\mathcal{S}\left( {\theta_t} \right) \right),
\end{split}
\end{equation}
where $\theta_0$ represents the initial parameters of the relay model, sampled from a specific distribution ${\Theta}$. The expectation on $\theta_0$ aims to improve the robustness of ${\mathcal{S}}$ to different parameter initialization \cite{lei_comprehensive_2024}. $\mathcal{D} \left( \cdot,\cdot \right) $ is the distance measurement. 
$\operatorname{opt}\left(\cdot\right)$ is the model parameter optimizer and the parameters of the relay model are updated only on $\mathcal{S}$.
The structure of condensed graph $\mathbf{A}'$ is parameterized by similarity in node features and modeled by a multi-layer perceptron (MLP) as ${\bf A}' = \text{MLP}({\bf X'})$.

\noindent \textbf{Limited distribution generalization of GC.} The optimization objective of GC aims to align the performance of GNNs trained on the condensed graph closely with those trained on the original graph. 
However, such objective is solely performance-driven and takes into account merely a snapshot of the original graph's distribution. Consequently, GNNs trained on the condensed graph are restricted to adapting to the condensed distribution and struggle to accommodate the dynamic distribution changes encountered during deployment.

\noindent \textbf{Intricate optimization in GC.}
The optimization procedure of GC is notably intricate, owing to two primary factors.
Firstly, the optimization objective in Eq. (\ref{eq_gmloss}) necessitates simultaneous updates of the relay model $f_{\theta}$ and the condensed graph ${\mathcal{S}}$. To tackle this demand, GC employs a nested loop optimization strategy, updating $f_{\theta}$ within the inner loop and optimizing ${\mathcal{S}}$ in the outer loop. 
Nevertheless, this bi-level optimization approach not only intensifies the computational demands but also introduces additional hyper-parameters, thus adding complexity to the optimization process and impeding the attainment of optimal solutions.
Secondly, the updating or periodic initialization of the relay model leads to the repetitive encoding of the original graph throughout the condensation process. 
Due to the neighbor explosion problem, the computational load for encoding the expansive original graph is substantial, and the repeated execution of the encoding operation exacerbates the computational burden even further.
Therefore, these intricate optimizations in GC constrain the frequent updating of condensed graph and its adaptability in life-long graph learning scenarios.

\begin{figure*}[t]
\setlength{\abovecaptionskip}{0.2cm}
\centering
\includegraphics[width=0.9\linewidth]{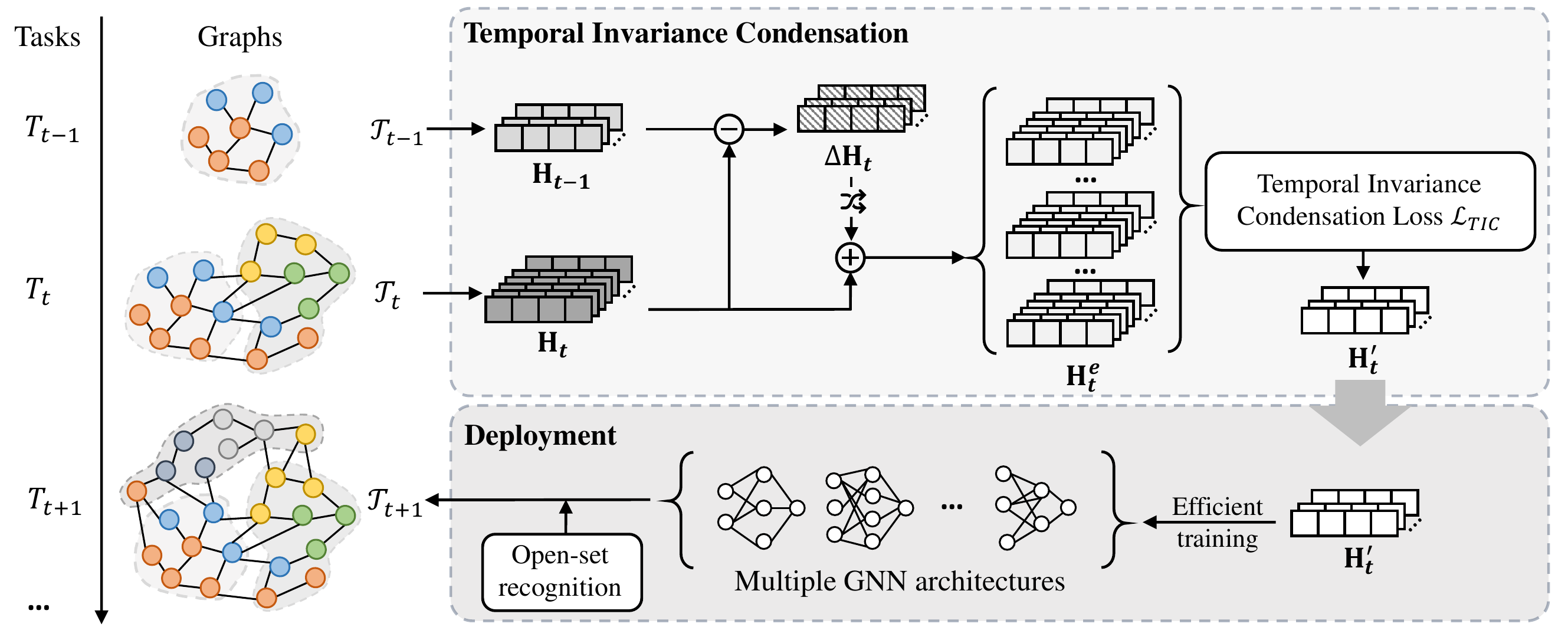}
\caption{The pipeline of OpenGC. The graph $\mathcal{T}_t$ and historic graph $\mathcal{T}_{t-1}$ are encoded by non-parametric convolution and embeddings are leveraged to construct temporal environments $\mathbf{H}^e_{t}$. The condensed graph embedding $\mathbf{H}'_t$ is generated according to temporal invariance condensation loss ${\mathcal{L}_{TIC}}$. In the deployment stage, the condensed graph is utilised to train multiple GNNs with various architectures, which are applied to sequential tasks $\mathcal{T}_{j\ge t}$.}
\label{fig_main}
\end{figure*}

\subsection{Problem Formulation}
\label{sec_formulation}
In the open-world graph scenario, we consider a sequential stream of node batches $\{V_1, V_2, ..., V_m\}$, where $m$ represents the total number of batches involved and each node batch accompanied by corresponding edges. 
These node batches are progressively accumulated and integrated into the existing graph over time to construct the tasks $\{T_1, T_2, ..., T_m\}$, as depicted in Figure \ref{fig_obs}. 
At task $T_t$, the snapshot graph $\mathcal{T}_t$ includes all $N_t$ nodes that have emerged and the graph expands as tasks progress, i.e., $\mathcal{T}_t \subset \mathcal{T}_{t+1}$. Correspondingly, the number of node class $C_t$ for task ${T}_t$ increases over time, satisfying  $C_t \le C_{t+1}$.

In this dynamic content, we consider the practical application scenario for both GC and GNN deployment.
During the GC phase at task  $T_t$, we initially annotate a subset of the newly added nodes in $\mathcal{T}_t$, addressing the continual integration of new classes.
Subsequently, the condensed graph ${\mathcal{S}_t}$ is generated for the large graph $\mathcal{T}_t$ and employed to train GNNs efficiently. 
In the deployment phase, GNNs trained on ${\mathcal{S}_t}$ are equipped with the open-set recognition method \cite{geng2020recent} for deployment on subsequent tasks $\mathcal{T}_{j\ge t}$.
This approach enables the recognition of nodes from new classes as the ``unknown class'', while nodes from observed classes in $\mathcal{T}_t$ are categorized into their appropriate classes. 
When significant structural changes occur, such as an influx of new class nodes or deteriorating GNN performance, the GC procedure is repeated to keep the condensed graph aligned with the latest original graph.

In this paper, our primary objective centers on enhancing both the efficiency and the generalization ability of GC across different distributions. The exploration of advanced open-set recognition methods falls outside the scope of this study.

%% file: 3met.tex
\section{Methodologies}
We hereby present our proposed open-world graph condensation (OpenGC). 
We begin with the foundation GC paradigm, which is an efficient GC approach to generate the condensed graph $\mathcal{S}_t$ for the static original graph $\mathcal{T}_t$.
Then we move on to temporal environment generation by exploring the structure-aware distribution shift. 
Finally, the generated environments are integrated to facilitate the temporal invariance condensation and the pipeline of OpenGC is depicted in Figure \ref{fig_main}.

\subsection{KRR-based Feature Condensation}

Based on our earlier discussions, the rationale for the intensive computational demand of the condensation process is the deficiency of the relay model $f= g \circ h$, which requires iterative optimization within the inner loop and repetitive encoding of the original graph. 
To address these challenges, we propose discarding the conventional relay model, which pairs GNNs with classifiers. Instead, we introduce a novel relay model that integrates non-parametric convolution with KRR to significantly improve the efficiency of static graph condensation. 
For the sake of simplicity, the task subscript $t$ is omitted in this subsection.

Specifically, we first transform the classification task into a regression problem by replacing the classification neural network $g$ with KRR. The corresponding loss function in Eq. (\ref{eq_iniloss}) is formulated as:
\begin{equation}
\label{eq_krrloss}
\begin{aligned}
&\mathcal{L}^{\mathcal{T}}(\theta) = \left \| {\bf Y}-h_{\theta}(\mathcal{T}){\bf W} \right \|^{2}+\lambda \left \|{\bf W} \right \|^{2},\\
&\mathcal{L}^{\mathcal{S}}(\theta) = \left \| {\bf Y}'-h_{\theta}(\mathcal{S} ){\bf W} \right \|^{2}+\lambda \left \|{\bf W} \right \|^{2},
\end{aligned}
\end{equation}
where ${\bf W}$ is the learnable matrix in KRR, $||\cdot||$ is the $L2$ norm, $\lambda$ is a constant, and $h_{\theta}(\cdot){\bf W}$ is the prediction of the labels.
Accordingly, the condensation objective in Eq. (\ref{eq_gmloss}) is substituted by Eq. (\ref{eq_krrgc}) to ensure that the regression model trained on $\mathcal{S}$ attains performance comparable to the one trained on the $\mathcal{T}$. 
\begin{equation}
\label{eq_krrgc}
\begin{aligned}
&\min_{\mathcal{S}}\mathbb{E}_{\theta \sim {\Theta}} \left \| {\bf Y}-h_{\theta}(\mathcal{T} ){\bf W}^{\mathcal{S}} \right \|^{2},\\
&s.t. \ \ {\bf W}^{\mathcal{S}} = \arg\min_{\bf W}\left \| {\bf Y}'-h_{\theta}(\mathcal{S}){\bf W} \right \|^{2}+\lambda \left \|{\bf W} \right \|^{2}.\\
\end{aligned}
\end{equation}

Consequently, KRR enables a closed-form, exact solution to the constraint above, eliminating iterative optimization of the classifier in the inner loop. The solution is calculated as:
\begin{equation}
\label{eq_solution}
 {\bf W}^{\mathcal{S}} =h_{\theta}(\mathcal{S})^{\mathsf{T}}\left(h_{\theta}(\mathcal{S})h_{\theta}(\mathcal{S})^{\mathsf{T}} + \lambda \mathbf{I}\right)^{-1} {\bf Y}',
\end{equation}
where $\mathbf{I}$ is the identity matrix. The reduction in graph size leads to an inversion matrix of dimensions $N' \times N'$, ensuring the computation remains efficient.

In addition to the classifier $g$, the graph encoder $h_{\theta}(\cdot)$ in conventional GC leverages GNNs that follow the message-passing paradigm \cite{gilmer2017neural}, involving the stacking of numerous propagation and transformation layers and leading to the iterative encoding issue. 
To alleviate this problem, we decompose the propagation and transformation processes within $h_{\theta}(\cdot)$, and then employ non-parametric graph convolution \cite{wu2019simplifying} to encode the original graph during the pre-processing stage. The $K$ layers convolution of the original graph and the condensed graph is calculated as:
\begin{equation}
\mathbf{H} = \hat{\mathbf{A}}^{K}\mathbf{X}, \; \; \; \; \; \mathbf{H}' = \hat{\mathbf{A}}'^{K}\mathbf{X}',
\label{eq_sgc}
\end{equation}
where $\hat{\mathbf{A}}'$ is the normalized adjacency matrix of $\mathbf{A}'$. Then, the encoder $h_{\theta}(\cdot)$ is constructed as:
\begin{equation}
\begin{aligned}
h_{\theta} \left(\mathcal{T} \right)=\phi_{\theta}(\mathbf{H}),\; \; \; \;  h_{\theta} \left(\mathcal{S} \right)=\phi_{\theta}( \mathbf{H}'),\\
\end{aligned}
\label{eq_randomness}
\end{equation}
where $\phi_{\theta}(\cdot)$ is the transformation layer parameterized by ${\theta}$~\cite{loo2022efficient}.
Finally, we use the identity matrix as the predefined adjacency matrix as previous works \cite{zheng_structure_free_2023,zhao2023dataset}, i.e, $\mathbf{A}' = {\mathbf{I}}$, to further simplify the condensed graph modeling and condense the original graph structure in the node attributes.
The benefit of this design is two-fold. Firstly, the predefined structure eliminates the training of the adjacency matrix generator. Secondly, it circumvents the encoding of the condensed graph in the condensation procedure. When conducting the downstream tasks, the identity matrix is leveraged as the adjacency matrix to train various GNNs.

\subsection{Structure-aware Environment Generation}
The condensed graph serves as a data simulator that resembles a snapshot of the original graph's distribution, consequently limiting the ability of GNNs trained on it to handle distribution shifts.
To counteract this, we incorporate invariance learning into the GC process, aiming to maintain temporal invariant patterns within the condensed graph.
Given the unavailability of future graph structures and distributions, we refer to the historic graph and construct various temporal environments by simulating potential future distribution shifts.

Specifically, we calculate the residuals by comparing the current embeddings at task $T_t$ with the embeddings from the last task $T_{t-1}$. 
The residual for the node $i$ is calculated as: 
\begin{equation}
\Delta\mathbf{H}_{t, i}={\mathbf{H}_{t, i}}-\mathbf{H}_{t-1, i},
\label{eq_res}
\end{equation}
where ${\mathbf{H}_{t, i}}$ and $\mathbf{H}_{t-1, i}$ are embeddings of node $i$ at task $T_t$ and $T_{t-1}$, respectively.
As the graph evolves, the neighbors of node $i$ are extended, and the node embedding changes from $\mathbf{H}_{t-1, i}$ to ${\mathbf{H}_{t, i}}$ after graph convolution. Therefore, $\Delta\mathbf{H}_{t, i}$ formulates the added neighbors in convolution and indicates the structure-aware distribution shift in task $T_t$, which can be leveraged to augment other node embeddings. 
Then, we randomly select node $j$, which belongs to the same class as node $i$, and use its residual $\Delta\mathbf{H}_{t, j}$ to modify the embedding $\mathbf{H}_{t, i}$.
The motivation hinges on the assumption that nodes in the same class follow a similar distribution shift pattern\footnote{We use the drop edge and drop feature to construct the environments if no historic classes are available.}.
\begin{equation}
\hat{\mathbf{H}}_{t, i}={\mathbf{H}_{t, i}}+ \varepsilon_{t, i} \Delta\bar{\mathbf{H}}_{t, j},
\label{eq_future}
\end{equation}
where $\Delta\bar{\mathbf{H}}_{t, j}$ is the normalized residual and $\varepsilon_{t, i}$ controls the magnitude. 
To determine the value of $\varepsilon_{t, i}$, we take into account the characteristics of the target nodes $i$ and the node $j$ responsible for generating the residual.
Firstly, a higher weight is assigned to the target node $i$ if it has a lower degree, with the premise that nodes with fewer connections are more prone to being influenced, leading to a pronounced distribution shift upon the addition of new neighbor nodes. 
Moreover, the similarity between node $i$ and $j$ is assessed to promote a more significant distribution shift among similar nodes. Consequently, the magnitude is constructed as:
\begin{equation}
\varepsilon_{t, i} = \delta_i \times \text{cosine}({\mathbf{H}_{t, i}}, {\mathbf{H}}_{t-1, j}) \times \eta ,
\label{eq_weight}
\end{equation}
where $\eta$ is the hyper-parameter to control the base magnitude and $\text{cosine}(\cdot)$ measures the consine similarity. $\delta_i$ is the degree calibration term sampled from the Beta distribution: $\delta_i \sim \text{Beta}(c\times degree_i, 1)$, where $degree_i$ denotes the degree of node $i$, $c$ represents the calibration constant. This introduces the randomness for multiple environments generation and a higher $degree_i$ increases the probability of sampling a smaller $\delta_i$. 

Finally, we produce several environments $e\in\mathcal{E}$ for invariance condensation, with the embeddings for each environment denoted by ${\mathbf{H}}^{{e}}_t= [ \hat{\mathbf{H}}_{t, 1};...; \hat{\mathbf{H}}_{t, N_t} ]$.

\subsection{Temporal Invariance Condensation}
With the definition of efficient graph condensation and multiple environment representations, we represent the final objective of temporal invariance condensation. 

At task $T_t$, we first pre-compute the embeddings of the original graph $\mathbf{H}_t$ and various environments ${\mathbf{H}}^{{e}}_t$.
To enhance the stability of the optimization, we incorporate a learnable temperature $\tau$ to calibrate the prediction of KRR and substitute the regression loss with cross-entropy loss. 
The loss for the original graph embedding $\mathbf{H}_t$ and condensed graph embedding $\mathbf{H}'_t$ is expressed as:
\begin{equation}
\begin{aligned}
&{\mathcal{L}(\mathbf{H}_t, f_{\theta})} = \frac{1}{N_t}\sum_{i=1}^{N_t}\sum_{j=1}^{C_t}\mathbf{Y}_{i,j}\text{log} \left[f_{\theta}(\mathbf{H}_t)/ \tau\right]_{i,j},\\
&f_{\theta}(\mathbf{H}_t)=\phi_{\theta}(\mathbf{H}_t) {\bf W}^{\mathcal{S}},\\
&{\bf W}^{\mathcal{S}} =\phi_{\theta}(\mathbf{H}'_t)^{\mathsf{T}}\left(\phi_{\theta}(\mathbf{H}'_t)\phi_{\theta}(\mathbf{H}'_t)^{\mathsf{T}} + \lambda \mathbf{I}\right)^{-1} {\bf Y}'_t,
\end{aligned}
\label{eq_firstloss}
\end{equation}
where ${\bf Y}'_t$ is the condensed graph label at task $T_t$.
Subsequently, the invariant risk minimization (IRM) loss \cite{arjovsky2019invariant} is constructed to identify and capture the invariance patterns across multiple environments:
\begin{equation}
{\mathcal{L}_{IRM}} = \frac{1}{|\mathcal{E}|} \sum_{e\in\mathcal{E}}   {\mathcal{L}({\mathbf{H}}^{{e}}_t, f_{\theta})}+    \gamma  { \left \|\nabla_{w|w=1}\mathcal{L}({\mathbf{H}}^{{e}}_t, wf_{\theta}) \right \|}^{2}.
\label{eq_secondloss}
\end{equation}
where $\gamma$ is the hyper-parameters to weight the loss. This constraint ensures uniform optimization results across all environments, thereby capturing the invariance patterns amidst various structural changes. 
Finally, the condensed graph is optimized according to the objective:
\begin{equation}
{\mathcal{L}_{TIC}} = \mathbb{E}_{\theta \sim {\Theta}} \left[ {\mathcal{L}}(\mathbf{H}_t, f_{\theta}) + \alpha {\mathcal{L}_{IRM}}\right],
\label{eq_overallloss}
\end{equation}
where $\alpha$ regulates the balance between original graph information and invariance information.

For a thorough comprehension of OpenGC, the summary of the training procedure and detailed time complexity analysis of our condensed data are presented in Appendix \ref{sec_al} and Appendix \ref{sec_com}, respectively.

%% file: 4exp.tex
\section{Experiments}

\begin{table*}[t]
\setlength{\abovecaptionskip}{0.1cm}
\centering
\caption{The comparison of mAP (\%) for different graph condensation methods on real-world datasets. Whole indicates the GNN is trained on the large original graph. Openmax and Softmax are deployed for open-set recognition in the testing stage.}
\label{tab_acc}
\resizebox{\textwidth}{!}{
\begin{tabular}{l|r|lllll|lllll}
\toprule
\multicolumn{1}{c|}{\multirow{2}{*}{}} & \multicolumn{1}{c|}{\multirow{2}{*}{\begin{tabular}[c]{@{}c@{}}Compress \\ ratio\end{tabular}}} & \multicolumn{5}{c|}{Openmax}                                                    & \multicolumn{5}{c}{Softmax}                                                     \\ \cline{3-12} 
\multicolumn{1}{c|}{}                  & \multicolumn{1}{c|}{}                                                                           & GCond      & GCDM       & SFGC       & OpenGC     & Whole                       & GCond      & GCDM       & SFGC       & OpenGC     & Whole                       \\ \midrule
\multirow{2}{*}{Yelp}                  & 0.01                                                                                            & 43.36±0.19 & 43.57±0.73 & 43.90±0.37 & 46.14±0.90 & \multirow{2}{*}{49.16±0.39} & 42.89±0.48 & 42.94±0.41 & 43.18±0.11 & 45.98±0.53 & \multirow{2}{*}{48.91±0.25} \\
                                       & 0.02                                                                                            & 43.61±0.35 & 43.79±0.33 & 44.09±0.12 & 46.42±0.19 &                             & 43.81±0.20 & 43.74±0.19 & 44.17±0.11 & 46.18±0.12 &                             \\ \hline
\multirow{2}{*}{Taobao}                & 0.01                                                                                            & 66.09±0.19 & 66.32±0.21 & 67.02±0.36 & 67.91±0.41 & \multirow{2}{*}{70.65±0.59} & 68.15±0.56 & 68.15±0.43 & 68.61±0.33 & 69.81±0.54 & \multirow{2}{*}{71.82±0.70} \\
                                       & 0.02                                                                                            & 66.84±0.98 & 66.58±0.74 & 67.22±0.32 & 68.26±0.83 &                             & 68.32±0.41 & 68.52±0.13 & 68.83±0.16 & 70.07±0.48 &                             \\ \hline
\multirow{2}{*}{Flickr}                & 0.001                                                                                           & 35.74±0.74 & 35.84±0.53 & 36.73±0.58 & 38.19±0.20 & \multirow{2}{*}{44.96±0.38} & 35.65±0.80 & 35.53±0.54 & 36.09±0.14 & 38.11±0.82 & \multirow{2}{*}{44.63±0.02} \\
                                       & 0.002                                                                                           & 36.58±0.95 & 36.20±0.53 & 36.97±0.14 & 39.06±0.66 &                             & 35.95±0.47 & 36.24±0.30 & 36.82±0.23 & 38.93±0.15 &                             \\ \hline
\multirow{2}{*}{Coauthor}              & 0.01                                                                                            & 68.96±0.30 & 68.91±0.28 & 69.27±0.31 & 69.94±0.38 & \multirow{2}{*}{72.44±0.16} & 69.31±0.17 & 71.48±0.33 & 71.63±0.83 & 73.00±0.60 & \multirow{2}{*}{74.44±0.15} \\
                                       & 0.02                                                                                            & 69.63±0.43 & 69.54±0.40 & 69.74±0.17 & 70.38±0.43 &                             & 71.35±0.44 & 71.53±0.34 & 71.94±0.15 & 73.32±0.92 &                             \\ \bottomrule
\end{tabular}}
\end{table*}

\begin{figure*}[t]
\setlength{\abovecaptionskip}{0.1cm}
\centering
\begin{minipage}[t]{0.23\linewidth}
\includegraphics[width=\linewidth]{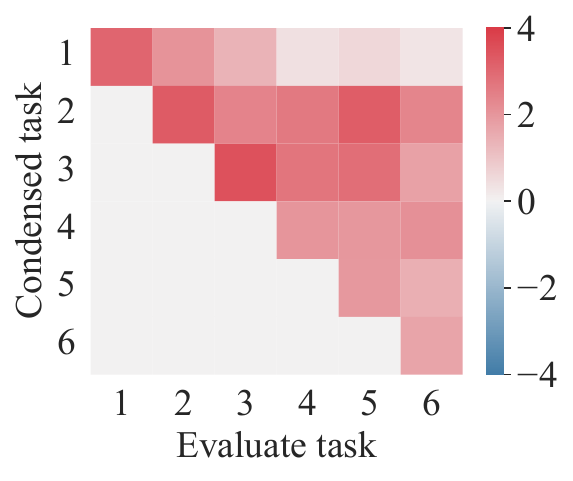}
\end{minipage}
\begin{minipage}[t]{0.23\linewidth}
\includegraphics[width=\linewidth]{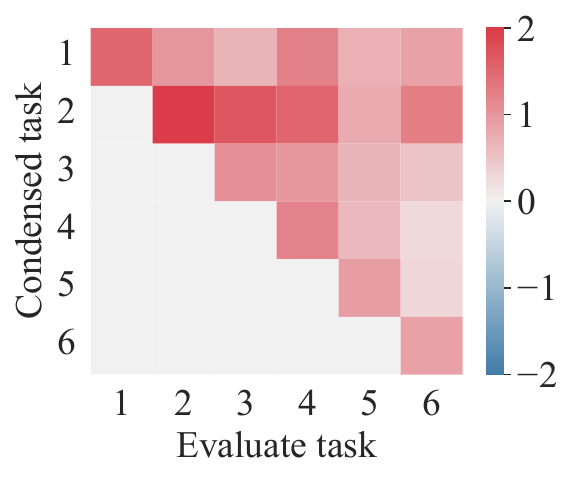}
\end{minipage}
\begin{minipage}[t]{0.23\linewidth}
\includegraphics[width=\linewidth]{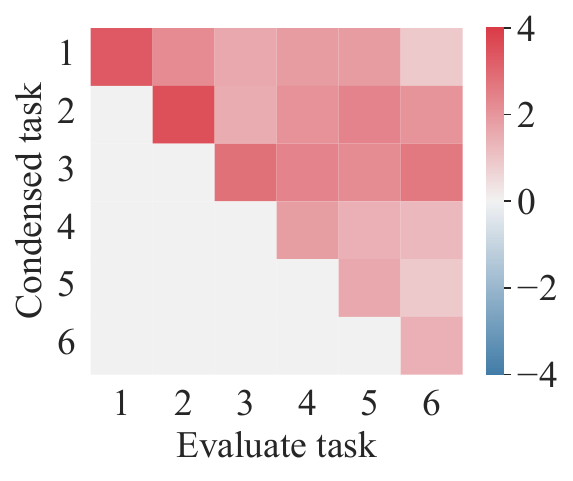}
\end{minipage}
\begin{minipage}[t]{0.23\linewidth}
\includegraphics[width=\linewidth]{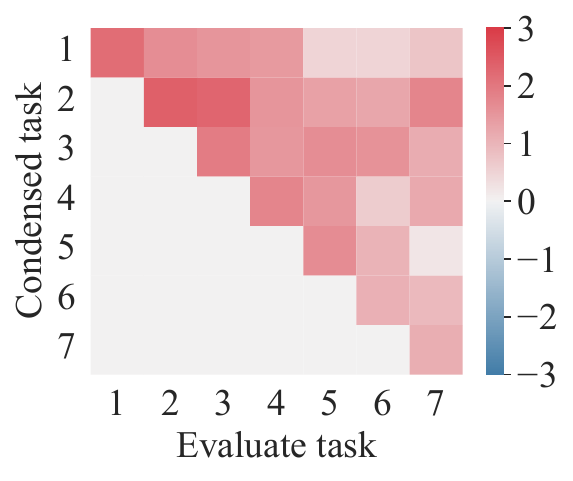}
\end{minipage}
\caption{The heatmap of the differences between performance matrix (\%) of OpenGC and SFGC on Yelp, Taobao, Flickr and Coauthor datasets (from left to right). Softmax is adopted and the compress ratios are 1\%, 1\%, 0.1\%, and 1\%, respectively.}
\label{fig_heat}
\end{figure*}

We design comprehensive experiments to validate the efficacy of our proposed OpenGC and explore the following research questions:
\textbf{Q1}: Compared to the SOTA GC methods, can OpenGC achieve better temporal generalization performance?
\textbf{Q2}: Can the OpenGC condense the graph faster than other GC approaches?
\textbf{Q3}: How do the different components, i.e., IRM and different constraints used in environment generation affect OpenGC?
\textbf{Q4}: Can the condensed graph generated by OpenGC generalize well to different GNN architectures? 
\textbf{Q5}: How do the different hyper-parameters affect the OpenGC?
\textbf{Q6}: What are the characteristics of the visualization of condensed nodes?

\subsection{Experimental Settings}
\label{exp_set}

\noindent\textbf{Datasets}. 
In dynamic real-world scenarios, newly added nodes often belong to both existing and novel classes and available datasets fail to fully capture this dynamic nature.
Consequently, we construct four real-world graph datasets to evaluate our proposed method: Yelp \footnote{https://www.yelp.com/dataset}, Taobao \footnote{https://tianchi.aliyun.com/dataset/dataDetail?dataId=9716}, Flickr \cite{DBLP:conf/iclr/ZengZSKP20} and Coauthor \cite{shchur2018pitfalls}. 
Following~\cite{feng2023towards}, we construct the Yelp and Taobao according to the natural timestamps to align with temporal sequences.
For the Flickr and Coauthor datasets, which do not include temporal information, we segment the data into tasks using randomly selected classes and nodes.
Without loss of generality, we guarantee that in subsequent tasks, newly added nodes not only introduce new classes but also belong to all classes observed in earlier tasks.
All datasets are randomly divided, with each task following a consistent split ratio: 60\% for training, 20\% for validation, and 20\% for testing. 
Moreover, the training, validation, and testing sets are continually expanded based on the sets from previous task. 
The detailed descriptions of datasets and tasks are provided in Appendix \ref{sec_data}.

\noindent\textbf{Baselines and Evaluation Settings}.
\label{sec_eval}
We compare our proposed methods to three SOTA GC methods:
(1) GCond \cite{jin2022graph}: the first GC method that employs the gradient matching and bi-level optimization techniques;
(2) GCDM \cite{zhao2023dataset}: an efficient GC method that utilizes distribution matching to align the class distributions; 
(3) SFGC \cite{zheng_structure_free_2023}: a structure-free GC method that pre-trains a large number of GNN models and utilizes trajectory matching to improve the optimization results.
To enable GNNs be capable of recognizing novel classes, we incorporate two open-set recognition techniques into each baseline: 
(1) Softmax \cite{geng2020recent}: it adds the softmax as the final output layer and sets the threshold to classify the low-confidence nodes as the ``unknown class''.
(2) Openmax \cite{bendale2016towards}: it assesses the probability of a node being an outlier via Weibull fitting and then calibrates the output logits, enabling the identification of novel classes with the predefined threshold.

We evaluate each GC method on evolving graphs containing a series of tasks 
 $\{T_1, T_2, ..., T_m\}$. 
 At task $T_t$, the graph $\mathcal{T}_t$ is condensed by GC methods, and the condensed graph is then used to train downstream GNNs.
The performance of these GNNs is evaluated across all subsequent tasks from $T_t$ to $T_m$. 
Therefore, an upper triangular performance matrix ${\mathbf M}\in{\mathbb{R}^{m\times m}}$ is maintained, where the element ${\mathbf M}_{i,j}$ represents the accuracy of a GNN trained on the condensed graph from task $T_i$ and tested on task $T_j$ $(i \le j)$. 
To quantify the overall performance of models trained on condensed graphs, we use the mean of average performance (mAP) as the evaluation metric as follows:
\begin{equation}
mAP = \frac{1}{m}\sum_{i=1}^{m} (\frac{1}{m-i+1}
  \sum_{j=i}^{m}\mathbf{M}_{i,j}).
\label{eq_map}
\end{equation}

For a fair comparison, we follow the conventional graph condensation \cite{jin2022graph} for the condensation setting. For all compared baselines, we employ SGC \cite{wu2019simplifying}, which uses non-parametric convolution, as the relay model and use the identity matrix as the adjacency matrix. The downstream GNN is GCN~\cite{DBLP:conf/iclr/KipfW17} unless specified otherwise. 

To assess the ability of GC to generalize across different GNN architectures, we conduct evaluations using a variety of models, including GCN~\cite{DBLP:conf/iclr/KipfW17}, SGC~\cite{wu2019simplifying}, GraphSAGE~\cite{hamilton2017inductive} and APPNP~\cite{klicpera2018predict}. We condense the original graph with $N_t$ nodes into a condensed graph with $N'_t$ nodes and the compress ratio is calculated by $N'_t/N_t$. We choose the compress ratio of Yelp, Taobao, and Coauthor to be \{1\%, 2\%\}. For the larger dataset Flickr, we choose the compress ratio as \{0.1\%, 0.2\%\}.

\noindent\textbf{Hyper-parameters and Implementation}. 
The hyper-parameters for baselines are configured as described in their respective papers, while others are determined through grid search on the validation set of the condensed task. 
For all datasets, a 2-layer graph convolution is employed, and the width of the hidden layer in $\phi_{\theta}$ is set to 1024. The regularization parameter $\lambda$ for KRR is fixed at 5e-3. The learning rate for the condensation process is determined through a search over the set \{1e-1, 1e-2, 1e-3, 1e-4\}. The parameters $\alpha$ and $\gamma$ are optimized from the set \{0.1, 0.3, 0.5, 0.7, 1\} to achieve an optimal balance between the losses. The intervention magnitude parameter $\eta$ is explored within the range \{0.01, 0.1, 1, 10, 100\} to identify the most appropriate level of intervention. The number of environments considered is varied from 1 to 5 to find the optimal setting. The calibration constant $c$ used in Beta distribution is set as 10. Finally, the number of training epochs is determined using the early stop to prevent overfitting.

Due to the absence of new class nodes in the validation set, we follow previous work \cite{bendale2016towards} to set the threshold for open-set recognition approaches, configuring it to exclude 10\% of the validation nodes as ``unknown class''. To eliminate randomness, we repeat each experiment 5 times and report the average test score and standard deviation.

We use the ADAM optimization algorithm to train all the models. The codes are written in Python 3.9 and the operating system is Ubuntu 16.0. We use Pytorch 1.12.1 on CUDA 11.4 to train models on GPU. All experiments are conducted on a machine with Intel(R) Xeon(R) CPUs (Gold 6128 @ 3.40GHz) and NVIDIA GeForce RTX 2080 Ti GPUs.

\begin{table*}[t]
\setlength{\abovecaptionskip}{0.1cm}
\centering
\caption{The comparison of condensation time (sec.) for different GC methods. The pre-processing time (Pre. time), average condensing time per 100 epochs (Avg. time), and total condensing time (Total time) are measured separately. The condensation target is the largest graph in the final task and the compress ratios are 1\%, 1\%, 0.1\%, and 1\% for 4 datasets respectively. ``<1'' indicates that the time duration is less than 1 second.}
\label{tab_time}
\resizebox{\textwidth}{!}
{\begin{tabular}{l|rrr|rrr|rrr|rrr}
\toprule
               & \multicolumn{3}{c|}{Yelp}                                                                       & \multicolumn{3}{c|}{Taobao}                                                                     & \multicolumn{3}{c|}{Flickr}                                                                     & \multicolumn{3}{c}{Coauthor}                                                                   \\ \hline
               & \multicolumn{1}{l}{Pre. time} & \multicolumn{1}{l}{Avg. time} & \multicolumn{1}{l|}{Total time} & \multicolumn{1}{l}{Pre. time} & \multicolumn{1}{l}{Avg. time} & \multicolumn{1}{l|}{Total time} & \multicolumn{1}{l}{Pre. time} & \multicolumn{1}{l}{Avg. time} & \multicolumn{1}{l|}{Total time} & \multicolumn{1}{l}{Pre. time} & \multicolumn{1}{l}{Avg. time} & \multicolumn{1}{l}{Total time} \\ \midrule
GCond          & <1                        & 6.92                          & 238.42                          & <1                          & 8.76                          & 175.04                          & <1                          & 5.97                          & 307.33                          & <1                         & 15.37                         & 260.39                         \\
GCDM           & <1                          & 3.08                          & 57.92                           & <1                          & 5.18                          & 72.15                           & <1                          & 3.82                          & 76.44                           & <1                          & 5.83                          & 116.47                         \\
SFGC           & 2125.24                       & 109.08                        & 2546.28                         & 3272.66                       & 100.52                        & 3272.66                         & 11959.14                      & 131.80                        & 1219.32                         & 3710.99                       & 124.72                        & 1325.07                        \\
OpenGC w/o IRM & <1                          & 1.70                          & 13.15                           & <1                       & 4.56                          & 19.22                           & <1                          & 2.84                          & 61.12                           & <1                         & 3.52                          & 28.28                          \\
OpenGC         & <1                          & 1.84                          & 17.65                           & <1                          & 6.13                          & 25.56                           & 4.51                          & 3.41                          & 78.36                           & 3.11                          & 4.61                          & 37.14                          \\ \bottomrule
\end{tabular}}
\end{table*}

\begin{table}[t]
\centering
\caption{The comparison of GNN's accuracy (\%) trained on the different condensed graphs. Whole indicates that GNNs are trained on the original graph. The condensation and evaluation target is the largest graph of the final task. The compress ratios are 1\%, 1\%, 0.1\% and 1\% for 4 datasets respectively.}
\resizebox{\linewidth}{!}
{\begin{tabular}{l|llll}
\toprule
               & Yelp       & Taobao     & Flickr     & Coauthor   \\ \midrule
Whole          & 52.06±0.55 & 80.20±0.06 & 53.49±0.22 & 93.63±0.14 \\
GCond          & 44.52±1.02 & 72.80±0.61 & 45.18±1.03 & 88.21±1.51 \\
GCDM           & 44.23±0.80 & 72.75±0.52 & 44.27±1.30 & 87.60±0.15 \\
SFGC           & 45.81±0.25 & 73.92±0.11 & 46.72±0.43 & 89.81±0.31 \\
OpenGC w/o IRM & 46.62±0.32 & 74.39±0.34 & 47.25±0.55 & 90.25±0.46 \\
OpenGC         & 47.52±0.46 & 74.81±0.39 & 48.15±0.65 & 90.93±0.49 \\ \bottomrule
\end{tabular}}
\label{tab_staaccall}
\end{table}

\subsection{Accuracy Comparison (Q1)}
We report the mAP for different GC methods with standard deviation in Table~\ref{tab_acc}. 
In this table, Whole indicates that GNNs are trained on the original graph, achieving the highest performance. However, it suffers from substantial computational costs due to the large scale of the original graph.
Compared to Whole, GC methods maintain similar performance levels even under extreme compress rate on Yelp, Taobao, and Coauthor datasets, confirming the effectiveness of GC.
On the Flickr dataset, the performance gap is larger compared to other datasets, attributing to the dataset's significant imbalance issue. When comparing the different GC methods, performance differences emerge. For example, GCond achieves comparable performances with GCDM across different datasets and compress ratios. 
Benefiting from the advanced trajectory matching strategy, SFGC achieves consistent performance improvement compared to GCond and GCDM. This is due that trajectory matching can provide more precise optimization guidance for GC procedure compared to gradient and distribution matching and significantly enhance the quality of condensed graphs. 
Nonetheless, all these methods only condense the snapshot of the original graph and preserve the static graph information in the condensed graph, which limits their performance under the dynamic graph scenarios.
Our proposed OpenGC consistently outperforms other baselines. Remarkably, with substantial compress rates, OpenGC achieves results comparable to the Whole on the Taobao and Coauthor under the Softmax setting. Furthermore, in Figure \ref{fig_heat}, we present a detailed heatmap of the differences between the performance matrix ${\mathbf M}$ of OpenGC and the strongest baseline SFGC. 
The observed pattern reveals a deep red coloration along the diagonal line, indicating a significant enhancement in performance on condensed tasks.
As the graph evolves, the color gradually lightens and the increment gradually weaken. This phenomenon is attributed to the increasing number of classes, which makes the task more complicated.
When comparing the different compress ratios, OpenGC guarantees superior GNN performance at lower compress ratios compared to the higher compress ratios achieved by other GC methods.
All these results underscore the efficacy of our proposed temporal invariance condensation and the superior optimization results from the exact solution provided by KRR.

\begin{table}[]
\setlength{\abovecaptionskip}{0.1cm}
\caption{Ablation study of OpenGC. The compress ratios are 1\%, 1\%, 0.1\% and 1\% respectively. The open-set recognition method is Softmax.}
\resizebox{\linewidth}{!}
{\begin{tabular}{l|llll}
\toprule
                 & Yelp       & Taobao     & Flickr     & Coauthor   \\ \midrule
OpenGC           & 45.98±0.53 & 69.81±0.54 & 38.11±0.82 & 73.00±0.60 \\
w/o IRM          & 44.25±0.28 & 69.31±0.15 & 36.30±0.80 & 71.78±0.11 \\
w/o TEG          & 44.99±0.16 & 69.47±0.64 & 36.54±0.63 & 72.06±0.40 \\
w/o degree       & 45.17±0.16 & 69.63±0.19 & 36.68±0.35 & 72.51±0.11 \\
w/o similarity   & 45.69±0.39 & 69.51±0.74 & 37.27±0.14 & 72.47±0.13 \\ \bottomrule
\end{tabular}}
\label{tab_abl}
\end{table}

\subsection{Condensation Time Comparison (Q2)}
For the sake of convenient comparison, we evaluate the GC methods by measuring the condensation time for the largest graph in the final task on each dataset. Besides our proposed method, we also assess the performance of OpenGC without the environment generation and invariant learning (``OpenGC w/o IRM'') and the results are presented in Table \ref{tab_time}. 
This assessment includes measuring the pre-processing time, the average condensation time per 100 epochs, and the total condensation time (pre-processing time excluded). The corresponding accuracy results and time complexity are reported in Table \ref{tab_staaccall} and Appendix \ref{sec_com}.
Firstly, the pre-processing time of GCond, GCDM, and OpenGC w/o IRM contains the time of non-parametric graph convolution in Eq. (\ref{eq_sgc}). OpenGC additionally involves the environment generation time.
In contrast, SFGC demands the training of hundreds of GNN models on the original graph to serve as teacher models and provide training trajectories. This step is computationally demanding and time-consuming, significantly exceeding the time taken for condensation. Moreover, training a large number of GNNs contradicts the core motivation behind GC, which aims for efficient training of multiple GNNs using condensed graphs.
The average condensing time per 100 epochs reflects the complexity of the condensation process. GCond utilizes gradient matching to align the original graph and condensed graph, necessitating the additional back-propagation steps for gradient computation with respect to model parameters. In contrast, GCDM alters the optimization objective and leverages the distribution matching, eliminating the gradient calculation and achieving a faster condensation procedure. Nonetheless, GCDM still engages in nested loop optimization for relay model updates. Similarly, SFGC requires multiple updates to the relay model based on the condensed graph to resemble the training trajectories of the teacher model. 
Moreover, it utilises the graph neural tangent kernel to evaluate the quality of the condensed graph, which is computationally intensive and further increases the condensing time.
Our method exhibits the highest condensation speed. Compared to OpenGC without IRM, the time required for environment generation in OpenGC is minimal. For example, the pre-processing on Yelp and Taobao takes less than 1 second. Despite the larger feature dimensions of Flickr and Coauthor, the pre-processing time remains under 10 seconds. Although GCDM also demonstrates rapid condensation, its accuracy is considerably lower than our proposed method, as detailed in Table \ref{tab_staaccall}.

\subsection{Ablation Study (Q3)}
To validate the impact of individual components, OpenGC is tested with Softmax, selectively disabling certain components. We evaluate OpenGC in the following configurations: (1) without the IRM loss (``w/o IRM''); (2) without the temporal environment generation (``w/o TEG''), using edge and feature drop to randomly generate environments instead; (3) without the degree constraints (``w/o degree''); (4) without the similarity constraint (``w/o similarity''). 
The results of these settings are presented in Table \ref{tab_abl}.
The removal of IRM loss leads to a noticeable decline in accuracy, underscoring the importance of both temporal environment generation and the invariance constraint in improving the generalization capabilities of the condensed graph.
The significance of temporal environment generation is further highlighted by the performance drop in ``w/o TEG''. Although new environments are generated, ``w/o TEG'' does not simulate the graph expanding pattern, failing to predict the future graphs and leading to the sub-optimal results.
Further analysis is conducted on the degree and similarity constraints during temporal environment generation. The performance in these cases is better than ``w/o TEG'' but does not reach the levels achieved by OpenGC, reinforcing the importance of incorporating node characteristics to enhance the environment generation.

\begin{figure*}[ht]
\setlength{\abovecaptionskip}{0.1cm}
\centering
\begin{minipage}[t]{0.24\linewidth}
\centering
\includegraphics[width=\linewidth]{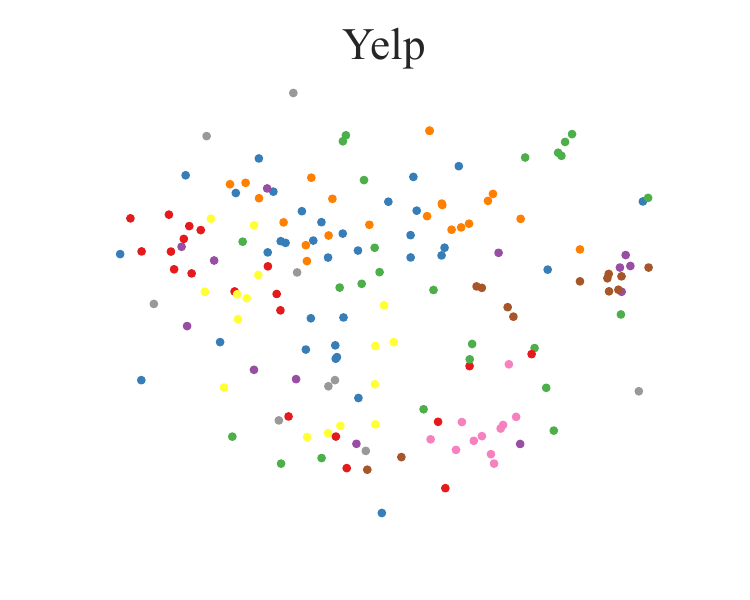}
\end{minipage}
\begin{minipage}[t]{0.24\linewidth}
\centering
\includegraphics[width=\linewidth]{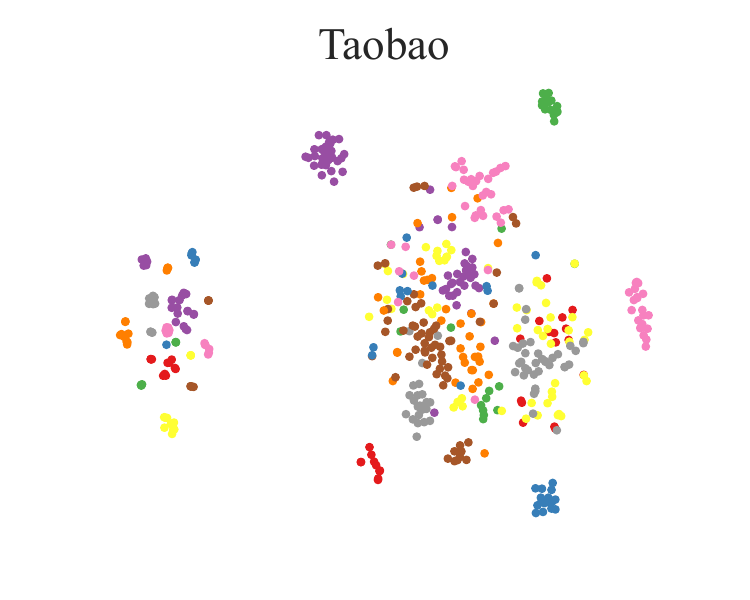}
\end{minipage}
\begin{minipage}[t]{0.24\linewidth}
\centering
\includegraphics[width=\linewidth]{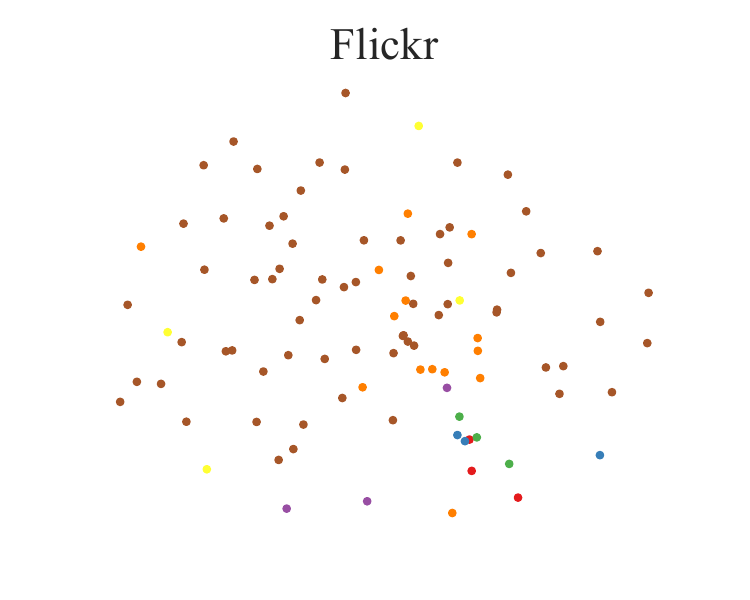}
\end{minipage}
\begin{minipage}[t]{0.24\linewidth}
\centering
\includegraphics[width=\linewidth]{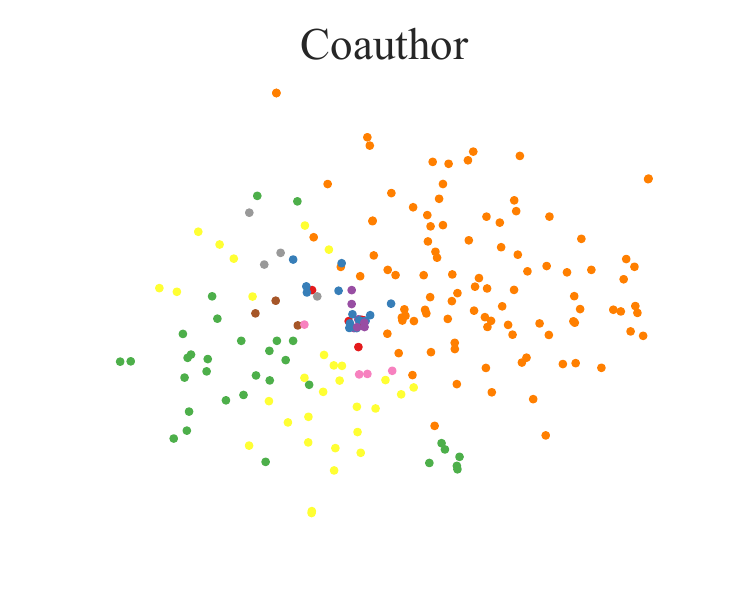}
\end{minipage}
\caption{The visualization of t-SNE on condensed graph by OpenGC.}
\label{fig_vis}
\end{figure*}

\begin{table}[t]
\setlength{\abovecaptionskip}{0.1cm}
\caption{Architecture generalizability of different graph condensation methods. The compress ratios are 1\%, 1\%, 0.1\%, and 1\% respectively. The open-set recognition method is Softmax.}
\resizebox{\linewidth}{!}
{\begin{tabular}{l|l|llll}
\toprule
                          & Method & GCN        & SGC        & GraphSAGE       & APPNP      \\ \midrule
\multirow{4}{*}{Yelp}     & GCond  & 42.89±0.48 & 42.41±0.49 & 41.02±0.44 & 42.01±0.35 \\
                          & GCDM   & 42.94±0.41 & 42.52±0.36 & 41.19±0.24 & 41.20±0.33 \\
                          & SFGC   & 43.18±0.11 & 43.96±0.79 & 43.00±0.18 & 43.54±0.35 \\
                          & OpenGC & 45.98±0.53 & 45.81±0.51 & 44.07±0.33 & 44.89±0.90 \\ \hline
\multirow{4}{*}{Taobao}   & GCond  & 68.15±0.56 & 68.51±0.11 & 70.03±0.32 & 67.33±0.05 \\
                          & GCDM   & 68.15±0.43 & 69.56±0.29 & 70.13±0.31 & 67.11±0.53 \\
                          & SFGC   & 68.61±0.33 & 69.77±0.58 & 70.92±0.26 & 68.01±0.11 \\
                          & OpenGC & 69.81±0.54 & 70.67±0.32 & 71.37±0.41 & 68.90±0.51 \\ \hline
\multirow{4}{*}{Flickr}   & GCond  & 35.65±0.80 & 35.57±0.15 & 33.80±0.31 & 34.55±0.04 \\
                          & GCDM   & 35.53±0.54 & 35.30±0.44 & 32.35±0.43 & 33.91±0.65 \\
                          & SFGC   & 36.09±0.14 & 36.39±0.16 & 33.82±0.50 & 34.57±0.15 \\
                          & OpenGC & 38.11±0.82 & 37.85±0.38 & 34.62±0.26 & 35.52±0.45 \\ \hline
\multirow{4}{*}{Coauthor} & GCond  & 69.31±0.17 & 70.17±0.08 & 68.07±0.55 & 68.86±0.31 \\
                          & GCDM   & 71.48±0.33 & 71.53±0.23 & 67.95±0.35 & 68.29±0.70 \\
                          & SFGC   & 71.63±0.83 & 71.18±0.11 & 69.03±0.23 & 69.51±0.21 \\
                          & OpenGC & 73.00±0.60 & 72.72±0.42 & 69.96±0.16 & 70.63±0.12 \\ \bottomrule
\end{tabular}}
\label{tab_crossarc}
\end{table}

\subsection{Generalizability for GNN Architectures (Q4)}
A critical attribute of GC is its ability to generalize across different GNN architectures, making the condensed graph versatile for training various GNN models in downstream tasks.
Therefore, we evaluate different GNN models on the condensed graph, including GCN, SGC, GraphSAGE, and APPNP.
These models are then applied to subsequent tasks and the mAP of different datasets are presented in Table \ref{tab_crossarc}.
According to the results, all evaluated GNN models were effectively trained using GC methods, achieving comparable levels of performance.
In detail, GCN and SGC exhibited superior performance due to these two models utilise the same convolution kernel as the relay model. GraphSAGE performed exceptionally well on the Taobao dataset, suggesting a particular compatibility with this datasets.
Moreover, different condensation methods exhibited varied performances, with our proposed method consistently maintaining the best results across various architectures.

\subsection{Hyper-parameter Sensitivity Analysis (Q5)}
\begin{figure}[t]
\setlength{\abovecaptionskip}{0.1cm}
\centering
\begin{minipage}[t]{0.43\linewidth}
\centering
\includegraphics[width=\linewidth]{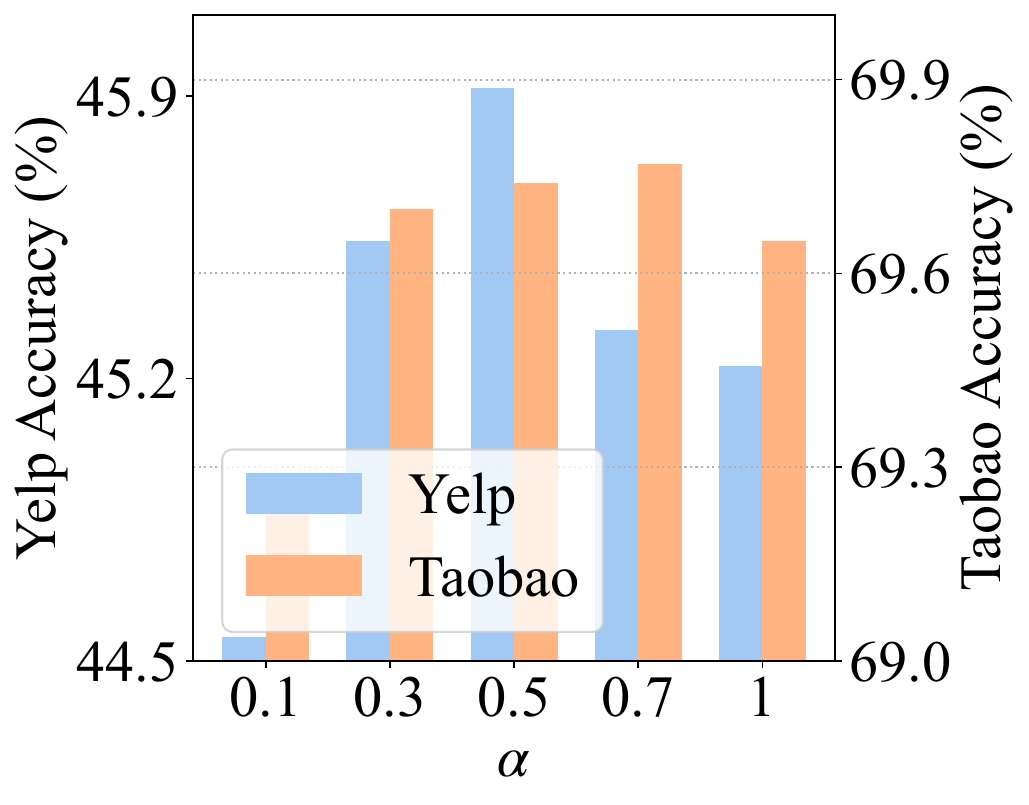}
\end{minipage}
\begin{minipage}[t]{0.43\linewidth}
\centering
\includegraphics[width=\linewidth]{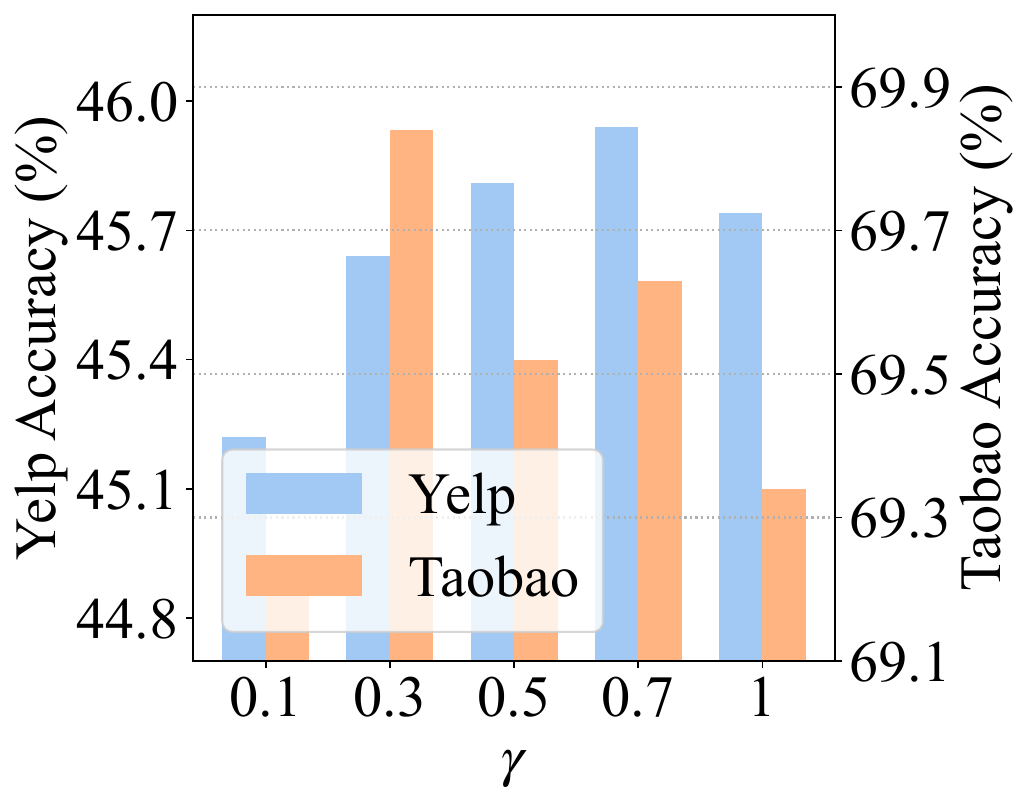}
\end{minipage}

\begin{minipage}[t]{0.43\linewidth}
\centering
\includegraphics[width=\linewidth]{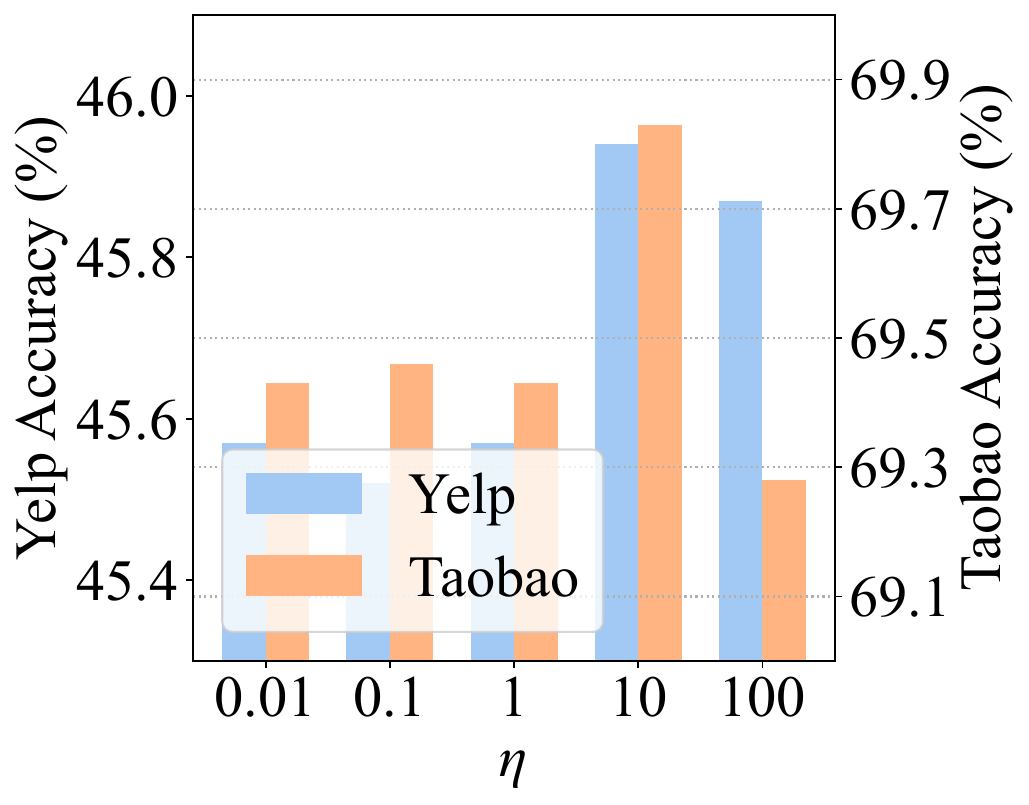}
\end{minipage}
\begin{minipage}[t]{0.43\linewidth}
\centering
\includegraphics[width=\linewidth]{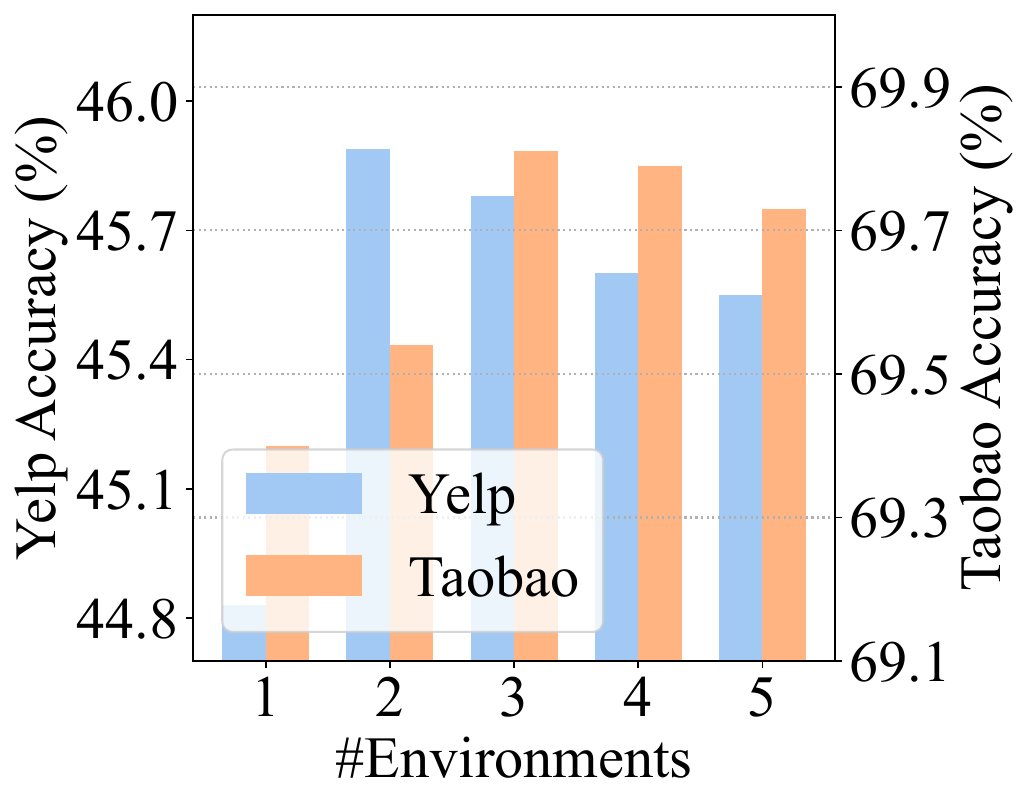}
\end{minipage}
\caption{The hyper-parameter sensitivity analysis.}
\label{fig_hyper}
\end{figure}

The hyper-parameters $\alpha$ and $\gamma$ are leveraged to control the impact of the IRM loss during training, while $\eta$ and the number of environments contribute to the extent of augmentations.
In Figure \ref{fig_hyper}, we present the mAP performances w.r.t. different values of $\alpha$, $\gamma$, $\eta$ and the number of environments, respectively. 
For $\alpha$, we observe that values in the range of 0.3 to 0.7 yield the best performance. Higher values may excessively weigh the IRM loss, potentially compromising the preservation of original graph information.
Similarly, selecting an optimal $\gamma$ value is crucial for maximizing performance across different datasets.
Regarding $\eta$, a value of 10 ensures a magnitude balance of the residual to the original embeddings. 
Finally, the increase of the number of environments can enhance performance and it should be controlled to avoid introducing excessive noise.

\subsection{Visualization (Q6)}
\label{sec_vis}

Figure \ref{fig_vis} presents the t-SNE visualization of node features in the condensed graph produced by our proposed OpenGC at the minimum compress ratios.
Although we eliminate the modeling the adjacency matrix and use the identity matrix instead, the condensed node features exhibit a well-clustered pattern, even on the highly imbalanced dataset Flickr. This suggests that the structure and class information are effectively preserved within the node features, enabling the training of GNNs for classification tasks in the absence of adjacency matrix.

%% file: 5con.tex
\section{Conclusion}
In this paper, we present open-world graph condensation (OpenGC), a robust graph condensation approach that enhances the temporal generalization capabilities of condensed graph. OpenGC exploits the structure-aware distribution shift in the evolving graph and extracts invariant features in the original graph for temporal invariance condensation.
Moreover, OpenGC optimizes the condensation procedure by combining kernel ridge regression and non-parametric graph convolution, successfully accelerating the condensation progress.
Benefiting from the superiority of the generalization capacity of condensed graphs and efficient optimization procedure, OpenGC not only enhances the GNNs' ability to handle the dynamic distribution change in real-world scenarios but also expedites the condensed graph updating in life-long graph learning.

%% file: ack.tex
\section*{Acknowledgment}
This work is supported by Australian Research Council under the streams of Future Fellowship (Grant No. FT210100624), Discovery Early Career Researcher Award (Grants No. DE230101033), Discovery Project (Grants No.DP240101108 and No.DP240101814). 

%% file: 6app.tex
\appendix
\section{Appendix}

\subsection{Algorithm of OpenGC}
\label{sec_al}
The detailed algorithm of OpenGC is shown in Algorithm \ref{al}. In pre-processing stage, we first encode both the original graph at task $T_t$ and $T_{t-1}$ by using non-parametric graph convolution. Then, multiple environments are generated according to calculated embeddings.
In condensation stage, we sample a initialization ${\theta}$ for $\phi_{\theta}(\cdot)$ from the distribution ${\Theta}$ . Then, the KRR solution ${\bf W}^{\mathcal{S}}$ is calculated according to the condensed graph.  Finally, the embeddings of constructed environments and original  graph are condensed via loss $\mathcal{L}_{TIC}$.

\subsection{Time Complexity Analysis}
\label{sec_com}
We show the detailed time complexity of OpenGC and compared baselines in Table \ref{tab_com}. Time complexities for both the pre-processing and condensing phases are assessed separately and the condensing procedure is further divided into the forward propagation, process execution, loss calculation, condensed graph updating, and relay model updating. The relay model for all methods is $K$ layer SGC incorporating a linear layer with the hidden dimension denoted by $b$. 
For the original graph, $N$, $E$, $d$ and $C$ are the number of nodes, edges, feature dimensions, and classes, respectively. The number of nodes in the condensed graph is represented by $N'$ and the pre-defined adjacency matrix is utilized across all methods.

The pre-processing stage for GCond and GCDM incorporates non-parametric graph convolution. SFGC entails the training of hundreds of teacher models and the quantity is denoted by $Z$. OpenGC incorporates environment generation during pre-processing phase, introducing an additional time complexity of $N'd$.

Due to the different optimization strategies utilized in GC methods, we decompose the condensing procedure into 5 stages. The process execution stage varies between methods, involving different operations specific to each method.
Specifically, GCond's process entails calculating the gradient w.r.t the relay model parameters twice. GCDM's procedure involves computing the class representation. SFGC necessitates updating the relay model on the condensed graph $q$ times to generate trajectories during each iteration.
OpenGC introduced the KRR for the closed-form solution, with the time complexity being $\mathcal{O} \left(N'^3+Nbc \right)$. Considering $N' \ll N$ and eliminating the relay model updating, our proposed method achieves a more efficient condensation procedure compared to other baselines.

\begin{algorithm}[ht]
\SetAlgoVlined
\textbf{Input:} Original graph at task $T_t$ and historic task $T_{t-1}$: $\mathcal{T}_t$ and $\mathcal{T}_{t-1}$\\
\textbf{Output:} Condensed graph $\mathcal{S}_t=\{{\bf I}, {\bf X}'_t\}$\\
$\rhd$  Pre-processing  \\
Compute embedding $\mathbf{H}_t$ and $\mathbf{H}_{t-1}$ with Eq. (\ref{eq_sgc})\\
Generate environments $\mathbf{H}^e_t$ with Eq. (\ref{eq_future})\\
$\rhd$  Graph condensation\\
Initialize ${\bf X}_t'$ and ${\bf Y}_t'$ \\
\For{$l=1,\ldots,L$}  
{
Initialize ${\theta}\sim {\Theta}$\\
Compute embedding $\mathbf{H}'_{t}$ of $\mathcal{S}_t$ with Eq. (\ref{eq_sgc})\\
Compute ${\bf W}^{\mathcal{S}}$ with Eq. (\ref{eq_firstloss}) \\
Compute $\mathcal{L}_{{TIC}}$ with Eq. (\ref{eq_overallloss})\\
${\bf X}'_t \leftarrow {\bf X}'_t -\mu \nabla_{{\bf X}'_t} \mathcal{L}_{{TIC}}$\\
}
\textbf{Return:} Condensed graph $\mathcal{S}_t$\\
\caption{The optimization framework of OpenGC}
\label{al}
\end{algorithm}

\begin{table*}[h]
\setlength{\abovecaptionskip}{0.1cm}
\caption{The comparison of the time complexity. The process in condensing procedure varies for different GC methods. GCond includes the gradient calculation. GCDM includes class representation calculation. OpenGC includes the calculation of KRR.}
\resizebox{0.65\linewidth}{!}
{\begin{tabular}{l|l|lllll}
\toprule
       & \multirow{2}{*}{Pre-processing} & \multicolumn{5}{c}{Condensing procedure}   \\ \cline{3-7} 
       &       & Forward & Process      & Loss & Update $\mathcal{S}$& Update $f$\\ \midrule
GCond  & $\mathcal{O} \left(KEd \right)$ & $\mathcal{O} \left( \left(N+N' \right)db \right)$ & $\mathcal{O} \left(db \right)$           & $\mathcal{O} \left(db \right)$  & $\mathcal{O} \left(N'd \right)$           & $\mathcal{O} \left(t_{in}db \right)$      \\
GCDM   & $\mathcal{O} \left(KEd \right)$ & $\mathcal{O} \left( \left(N+N' \right)db \right)$ & $\mathcal{O} \left( \left(N+N' \right)b \right)$      & $\mathcal{O} \left(Cb \right) $ & $\mathcal{O} \left(N'd \right)$           & $\mathcal{O} \left(t_{in}db \right)$\\
SFGC   & $\mathcal{O} \left(Z(KEd+Ndb) \right)$& $\mathcal{O} \left(qN'db \right)$    & N/A& $\mathcal{O} \left(db \right) $ & $\mathcal{O} \left(N'd \right)$           & $\mathcal{O} \left(qdb \right)$\\
OpenGC & $\mathcal{O} \left(KEd+N'd \right)$          & $\mathcal{O} \left( \left(N+N' \right)db \right)$ & $\mathcal{O} \left(N'^3+NCb \right)$ & $\mathcal{O} \left(Cb \right)$  & $\mathcal{O} \left(N'd \right)$  & N/A \\ \bottomrule
\end{tabular}}
\label{tab_com}
\end{table*}

\begin{table*}[t]
\setlength{\abovecaptionskip}{0.1cm}
\caption{Dataset statistics.}
\resizebox{0.55\textwidth}{!}
{
\begin{tabular}{l|rrrlrrl}
\toprule
         & \multicolumn{1}{l}{\#Nodes} & \multicolumn{1}{l}{\#Edges} & \multicolumn{1}{l}{\begin{tabular}[c]{@{}l@{}}\# Total \\ classes\end{tabular}} & \# Timespan & \multicolumn{1}{l}{\# Tasks} & \multicolumn{1}{l}{\begin{tabular}[c]{@{}l@{}}\# Classes \\ per task\end{tabular}} & \begin{tabular}[c]{@{}l@{}}\# Timespan \\ per task\end{tabular} \\ \midrule
Yelp     & 12,853                       & 179,612                      & 12                                   & 6 years     & 6                            & 2                                       & 1 year               \\
Taobao   & 51,358                       & 364,010                      & 18                                   & 6 days      & 6                            & 3                                       & 1 day                \\
Flickr   & 89,250                       & 899,756                      & 7                                    & N/A           & 6                            & 1                                       & N/A                   \\
Coauthor & 18,333                       & 163,788                      & 15                                   & N/A           & 7                            & 2                                       & N/A                    \\ \bottomrule
\end{tabular}}
\label{tab_datastat}
\end{table*}

\begin{table*}[t]
\setlength{\abovecaptionskip}{0.1cm}
\caption{The node, edge, and class distribution of different tasks.}
\resizebox{0.58\textwidth}{!}
{
\begin{tabular}{l|l|rrrrrrr}
\toprule
Dataset                   & Task    & 1    & 2     & 3     & 4      & 5      & 6      & 7             \\  \midrule
\multirow{3}{*}{Yelp}     & \#node  & 1,623 & 3,504  & 6,491  & 8,735   & 10,259  & 12,853  &  N/A       \\
                          & \#edge & 7,682 & 22,306 & 65,662 & 109,298 & 133,148 & 179,612 &  N/A       \\ 
                          & \#class & 2 & 4 & 6 & 8 & 10 & 12 &  N/A       \\ \hline
\multirow{3}{*}{Taobao}   & \#node  & 1,817 & 6,553  & 12,151 & 15,979  & 27,399  & 51,358  &  N/A       \\
                          & \#edge & 6,726 & 23,686 & 51,480 & 71,902  & 137,376 & 364,010 &  N/A       \\ 
                          & \#class & 3 & 6 & 9 & 12 & 15 & 18 &  N/A       \\ \hline
\multirow{3}{*}{Flickr}   & \#node  & 2,294 & 5,870  & 10,671 & 23,127  & 37,322  & 89,250  &  N/A       \\
                          & \#edge & 5,742 & 19,496 & 41,184 & 156,654 & 268,810 & 899,756 &  N/A       \\ 
                          & \#class &2 & 3 & 4 & 5 & 6 & 7 &  N/A      \\ \hline
\multirow{3}{*}{Coauthor} & \#node  & 459   & 1,221  & 2,495  & 4,193   & 6,411   & 9,855   & 18,333    \\
                          & \#edge & 418   & 2,310  & 7,052  & 16,194  & 32,988  & 64,480  & 163,788 \\ 
                          & \#class & 3 & 5 & 7 & 9 & 11 & 13 & 15      \\ \bottomrule
\end{tabular}}
\label{tab_taskstat}
\end{table*}

\subsection{Dataset Details}
\label{sec_data}
We follow~\cite{feng2023towards} to process Yelp and Taobao to construct the open-world graph datasets.

Yelp \footnote{https://www.yelp.com/dataset} is a large business review website where people can upload their reviews for commenting business, and find their interested business by others’ reviews. According to reviews, we construct a business-to-business temporal graph. Specifically, we take the data from 2016 to 2021, and treat the data in each year as a task, thus forming 6 tasks in total. In each year, we sample the 2 largest business categories as classes in each task. The newly added nodes in later tasks will cover all observed classes in former task classes. We regard each business as a node and set the business’s category as its node label. The temporal edge will be formed, once a user reviews the corresponding two businesses within a month. We initialize the feature representation for each node by averaging 300-dimensional GloVe word embeddings of all reviews for this business following the previous work \cite{hamilton2017inductive}.

Taobao \footnote{https://tianchi.aliyun.com/dataset/dataDetail?dataId=9716} is a large online shopping platform where items can be viewed and purchased by people online. For the Taobao dataset, we construct an item-to-item graph, in the same way as Yelp. The data in the Taobao dataset is a 6-day promotion season of Taobao in 2018. The data in each day is treated as a task and we sample 3 largest item categories in each task. 
We regard the items as nodes and take the categories of items as the node labels. The temporal edge will be built if a user purchases 2 corresponding items in the promotion season. We use the 128-dimensional embedding provided by the original dataset as the initial feature of the node.

Flickr \cite{DBLP:conf/iclr/ZengZSKP20} is an image network where each node in the graph represents one image. If two images share some common properties (e.g., same geographic location, same gallery, comments by the same user, etc.), an edge will be established
between these two images. All nodes are classified into 7 classes and node features are the 500-dimensional bag-of-word representations. We randomly choose 2 classes as the initial task and each subsequent task adds a new class. 

Coauthor \cite{shchur2018pitfalls} is a co-authorship graph. Nodes represent authors and are connected by an edge if they co-authored a paper. Node features represent paper keywords for each author’s papers, and class labels indicate the most active fields of study for each author. Similar to Flicker, we randomly choose 3 classes as the initial task and each subsequent task adds 2 new classes. 

The detailed dataset statistics and task distribution for each dataset are shown in Table \ref{tab_datastat} and Table \ref{tab_taskstat}, respectively.

\subsection{Related Work}

\noindent\textbf{Graph condensation.}
Graph condensation is designed to reduce GNN training costs through a data-centric perspective and most GC methods focus on improving the GNN accuracy trained on the condensed graph. 
For example, SFGC \cite{jin2022condensing} introduces trajectory matching in GC and proposes to align the long-term GNN learning behaviors between the original graph and the condensed graph.
GCEM \cite{liu_graph_2023} focuses on improving the performance of different GNN architectures trained on the condensed graph. It circumvents the conventional relay GNN and directly generates the eigenbasis for the condensed graph to preserve high-frequency information. 
Besides improving the quality of condensed graphs, GC has been widely used in various applications due to its excellent graph compression performance, including inference acceleration \cite{gao_graph_2023}, continual learning \cite{liu_cat_2023}, hyper-parameter/neural architecture search \cite{ding_faster_2022} and federated learning \cite{pan_fedgkd_2023}.
Although GC is developed on various applications, none of them focus on the evolution of graphs in real-world scenarios. Our proposed method is the first to explore this practical problem and contains its significance in GC deployment.

\noindent\textbf{Invariant learning for out-of-distribution generalization.}
The invariant learning approaches are proposed to reveal invariant relationships between the inputs and labels across different distributions while disregarding the variant spurious correlations. 
Therefore, numerous methods are developed to improve the out-of-distribution (OOD) generalization of models, which refers to the ability to achieve low error rates on unseen test distributions.
For example, Invariant Risk Minimization \cite{arjovsky2019invariant} improves the empirical risk minimization and includes a regularized objective enforcing simultaneous optimality of the classifier across all environments. 
Risk Extrapolation \cite{krueger2021out} encourages the equality of risks of the learned model across training environments to enhance the model sensitivity to different environments.
Recent works utilize invariant learning in OOD graph generalization problem \cite{wu2022discovering,zhu2021shift,yehudai2021local,liu2023flood}, and the most critical part of them is how to design invariant learning tasks and add proper regularization specified for extracting environment-invariant representations.
However, these methods are all model-centric and only concentrate on enhancing the GNN model to extract the invariant features. In contrast, our proposed method introduces invariant learning in the data-centric GC method, enabling the GNNs trained on the condensed graph all contain the OOD generalization.